\documentclass[runningheads]{llncs}

\usepackage{arxiv}

\long\def\comment#1{}
\usepackage{multirow}
\usepackage{amsmath}
\usepackage{amssymb}

\newcommand{\argmax}{\mathop{\mathrm{argmax}}}
\usepackage{graphicx}
\newcommand{\hochkomma}{$^{,}$}

\begin{document}

%\pretitle{}
\title{ProVe: A Pipeline for Automated Provenance Verification of Knowledge Graphs against Textual Sources}
\titlerunning{ProVe: A Pipeline for KG Provenance Verification}
%\subtitle{}

\author{Gabriel Amaral\inst{1}\orcidID{0000-0002-4482-5376} \and
{Odinaldo Rodrigues \inst{1}\orcidID{0000-0001-7823-1034}} \and
Elena Simperl \inst{1}\orcidID{0000-0003-1722-947X}}

\authorrunning{G. Amaral et al.}
\institute{King's College London, London WC2R 2LS, UK\\
\email{\{gabriel.amaral,odinaldo.rodrigues,elena.simperl\}@kcl.ac.uk}}

\maketitle

\begin{abstract}
%You need to decide whether you are selling a generic system, applied to a Wikidata dataset, or a Wikidata solution. At the moment, the introduction reads like here's a Wikidata problem and the solution requires novel research, whereas the rest, in particular Sections 3-4 read like: here's a system, we just happenned to evaluate it on this KG. You need to adjust the whole paper to sell one story, not two. Also, the Wikidata stats in the abstract and intro/rest of the paper are not consistent. Please use the same numbers. The terminology is inconsistent as well: is it triples or semantic relationships, plus the names of the steps/components in your system. Decide on a terminology (which you will reuse in your thesis) and stick to it in this paper

Knowledge Graphs are repositories of information that gather data from a multitude of domains and sources in the form of semantic triples, serving as a source of structured data for various crucial applications in the modern web landscape, from Wikipedia infoboxes to search engines. Such graphs mainly serve as secondary sources of information and depend on well-documented and verifiable provenance to ensure their trustworthiness and usability. However, their ability to systematically assess and assure the quality of this provenance, most crucially whether it properly supports the graph's information, relies mainly on manual processes that do not scale with size. ProVe aims at remedying this, consisting of a pipelined approach that automatically verifies whether a Knowledge Graph triple is supported by text extracted from its documented provenance. ProVe is intended to assist information curators and consists of four main steps involving rule-based methods and machine learning models: text extraction, triple verbalisation, sentence selection, and claim verification. ProVe is evaluated on a Wikidata dataset, achieving promising results overall and excellent performance on the binary classification task of detecting support from provenance, with $87.5\%$ accuracy and $82.9\%$ F1-macro on text-rich sources. The evaluation data and scripts used in this paper are available in GitHub and Figshare.%We believe ProVe to be a valuable asset in the upkeep of verifiability in Knowledge Graphs.

% 3-7 according to the semantic web journal so let's go with 5
\end{abstract}

%%%%%%%%%%% The article body starts:

%
%\newpage
\section{Introduction}
\label{sec:intro}
%1.5 page would be ideal
% it had 3 now it has 2

A Knowledge Graph (KG) is a type of knowledge base that stores information in the form of semantic triples formed by a subject, a predicate, and an object. KGs represent both real and abstract entities internally as labelled and uniquely identifiable entities, such as \textit{The Moon} or \textit{Happiness}, and can amass information from a multitude of domains and sources by connecting such entities amongst themselves or to literals through relationships, coded via uniquely identified predicates. KGs serve as sources of both human and machine-readable semantically structured data for various crucial applications in the modern web landscape, such as Wikipedia infoboxes, search engines results, voice-activated assistants, and information gathering projects~\cite{malyshev2018getting}.

Developed and maintained by ontology experts, data curators, and even anonymous volunteers, KGs have massively grown in size and adoption in the last decade, mainly as secondary sources of information. This means not storing new information, but taking it from authoritative and reliable sources which are explicitly referenced. As such, KGs depend on well-documented and verifiable provenance to ensure they are regarded as trustworthy and usable~\cite{zaveri2016quality}.

Processes to assess and assure the quality of information provenance are thus crucial to KGs, especially measuring and maintaining verifiability, i.e. the degree to which consumers of KG triples can attest these are truly supported by their sources~\cite{zaveri2016quality}. However, such processes are currently performed mostly manually, which does not scale with size. Manually ensuring high verifiability on vital KGs such as Wikidata and DBpedia is prohibitive due to their sheer size. 

ProVe (Provenance Verification) is proposed to assist data curators and editors in handling the upkeep of KG verifiability. It consists of an automated approach that leverages state-of-the-art Natural Language Processing (NLP) models, public datasets on data verbalisation and fact verification, as well as rule-based methods. %Given a KG triple, a reference is defined as an accessible portion of its provenance which should be sufficient to verify it, such as a web page, a book, or a uniquely identified entry in an external database. 
ProVe consists of a pipeline that aims at automatically verifying whether a KG triple is supported by a web page that is documented as its provenance. ProVe first extracts text passages from the triple's reference. Then, it verbalises the KG triple and ranks the extracted passages according to their relevance to the triple. The most relevant passages have their stances towards the KG triple determined (i.e. supporting, refuting, neither) and finally ProVe estimates whether the whole reference supports the triple.

This task is a specific application of Automated Fact Checking (AFC), also known as AFC on KGs. AFC is a currently well explored topic of research with several published papers, surveys, and datasets~\cite{thorne2018automated,zeng2021automated,guo2022survey,soleimani2020bert,malon2019team,lewis2020rag,hanselowski2018ukp,zhong2019dream,liu2019kgat,zhou2019gear,thorne2018fact,thorne2018fever,sathe2020automated}, and generally defined as the verification of a natural language claim by collecting and reasoning over evidence extracted from text documents or structured data sources. Both the verification verdict and the collected evidence are the main outputs. While general AFC takes a textual claim and a searchable evidence base as inputs, AFC on KGs takes a single KG triple and its documented provenance in the form of an external reference. %This introduces extra layers of complexity~\cite{huaman2020knowledge}. Firstly, KG triples rely on multiple labels and descriptions to be fully meaningful. Secondly, the correct labels and interpretations might depend on context or conventions. Thirdly, we can not rely on neighbouring triples and must depend only on a pre-defined source which might potentially not be useful. 

Approaches tackling AFC on KGs are very few, with the two only works of note in a similar direction, as far as is known by the authors, being DeFacto~\cite{lehmann2012defacto,gerber2015defacto} and its successor FactCheck~\cite{syed2018factcheck}. While they tackle this task mostly as defined above, they rely on a searchable document base instead of a given reference and judge triples on a true-false spectrum instead of verifiability. Like these few approaches, ProVe diverges from the general AFC framework and introduces a few different sub-tasks. Still, it makes use of the current state-of-the-art on those subtasks in common, being the first approach to tackle AFC on KGs with large pre-trained Language Models (LMs), which can be expanded to work in languages other than English and benefits from an Active Learning scenario.

ProVe is evaluated on an annotated dataset of Wikidata triples and their references, combining multiple types of properties and web domains. ProVe achieves promising results overall ($75\%$ accuracy and $68.1\%$ F1-macro) on classifying references as either supporting their triples or not, with an excellent performance on explicit and naturally written references ($87.5\%$ accuracy, $82.9\%$ F1-macro, $0.908$ AUC). Additionally, ProVe assesses passage relevance with a strong positive correlation ($0.5058$ Pearson's r) to human judgements.

In summary, this paper's main contributions are:
\begin{enumerate}
    \item A novel pipelined approach to evidence-based Automated Fact Checking on Knowledge Graphs based on large Language Models;
    \item A benchmarking dataset of Wikidata triples and references for Automated Fact Checking on Knowledge Graphs, covering a variety of information domains as well a balanced sample of diverse web domains;
    \item Novel crowdsourcing task designs that facilitate repeatable, quick, and large-scale collection of human annotations on passage relevance and textual entailment at good agreement levels.
\end{enumerate}

These contributions directly aid KG curators, editors, and researchers in improving KG provenance. Properly deployed, ProVe can do so in multiple ways. Firstly, by assisting the detection of verifiability issues in existing references, bringing them to the attention of humans. Secondly, given a triple and its reference, it can promote re-usability of the reference by verifying it against neighbouring triples. Finally, given a new KG triple entered by editors or suggested by KG completion processes, it can analyse and suggest references. 
The remainder of this paper is structured as follows. Section~\ref{sec:related} explores related work on KG data quality, mainly verifiability, as well as approaches to AFC on KGs. Section~\ref{sec:method} presents ProVe's formulation and covers each of its modules in detail. Section~\ref{sec:data} presents an evaluation dataset consisting of triple-reference pairs, including its generation and its annotation. Section~\ref{sec:eval} details the results of ProVe's evaluation. Finally, Section~\ref{sec:disc} delivers discussions around this work and final conclusions. All code and data used in this paper are available on Figshare~\footnote{\url{https://figshare.com/s/df0ec1c233ebd50817f4}} and GitHub.~\footnote{\url{https://anonymous.4open.science/r/RSP-F367/}}\hochkomma\footnote{\url{https://anonymous.4open.science/r/ClaimVerificationHIT-A04D}}
%
%\newpage
\section{Related Work}
\label{sec:related}
% was 6.5 pages, now is 4

ProVe attempts to solve the task of AFC on KGs, with the purpose of assisting data curators in improving the verifiability of KGs. Thus, to understand how ProVe approaches this task, it is important to first understand how the data quality dimension of verifiability is currently defined and measured in KGs, as well as how state-of-the-art approaches to general AFC and AFC on KGs tackle these tasks and how ProVe learns or differs from them.

% KG DATA QUALITY
\subsection{Verifiability in KGs}

In order to properly evaluate the degree to which ProVe adequately predicts verifiability, this dimension first needs to be well defined and a strategy needs to be established to measure it given an evaluation dataset. Verifiability in the context of KGs, whose information is mainly secondary, is defined as the degree to which consumers of KG triples can attest these are truly supported by their sources~\cite{zaveri2016quality}. It is an essential aspect of trustworthiness~\cite{zaveri2016quality,farber2018linked,piscopo2019we}, yet is amongst the least explored quality dimensions~\cite{zaveri2016quality,piscopo2019we}, with most measures carried superficially, unlike correctness or consistency~\cite{piscopo2019we,shenoy2022study,acosta2018detecting,kontokostas2013triplecheckmate,acosta2013crowdsourcing}.

For instance, Farber et al.~\cite{farber2018linked} measure verifiability only by considering whether any provenance is provided at all. Flouris et al.~\cite{flouris2012using} look deeper into sources' contents, but only verify specific and handcrafted irrationalities, such as a city being founded before it had citizens. Algorithmic indicators are not suited to directly measure verifiability, as sources are varied and natural language understanding is needed. As such, recent works~\cite{piscopo2017provenance,amaral2021assessing} measure KG verifiability through crowdsourced manual verification, giving crowdworkers direct access to triples and references. Crowdsourcing allows for more subjective and nuanced metrics to be implemented, as well as for natural text comprehension~\cite{xue2022knowledge,cao2020knowledge}. 

Thus, this paper employs crowdsourcing in order to measure verifiability metrics of individual triple-reference pairs. By comparing a pair's metrics with ProVe's outputs given said pair as input, ProVe and its components can be evaluated. Like similar crowdsourcing studies~\cite{piscopo2017provenance,amaral2021assessing}, multiple quality assurance techniques are implemented to ensure collected annotations are trustworthy~\cite{daniel2018quality}. To the best of the authors' knowledge, this is the first work to use crowdsourcing as a tool to measure the relevance and stance of references in regards to KG triples at levels varying from whole references to individual text passages.

\subsection{Automated Fact Checking on Knowledge Graphs}
\label{sec:related_afc}

\subsubsection{General AFC}

%AFC definition
Automated Fact Checking (AFC) is a topic of several works of research, datasets, and surveys~\cite{thorne2018automated,zeng2021automated,guo2022survey,soleimani2020bert,malon2019team,lewis2020rag,hanselowski2018ukp,zhong2019dream,liu2019kgat,zhou2019gear,thorne2018fact,thorne2018fever,sathe2020automated}. AFC is commonly defined in the Natural Language Processing (NLP) domain as a broader category of tasks and subtasks~\cite{thorne2018automated,zeng2021automated,guo2022survey} whose goal is to, given a textual claim and searchable document corpora as inputs, verify said claim's veracity or support by collecting and reasoning over evidence. Such evidence is extracted from the input document corpora and constitutes AFC's output alongside the claim's verdict. While a detailed exploration of individual AFC state-of-the-art approaches is out of this paper's scope, it is crucial to define their general framework in order to properly cover ProVe's architecture.

\begin{figure}[ht]
  \centering
  \vspace*{-15pt}
  \includegraphics[width=1\linewidth]{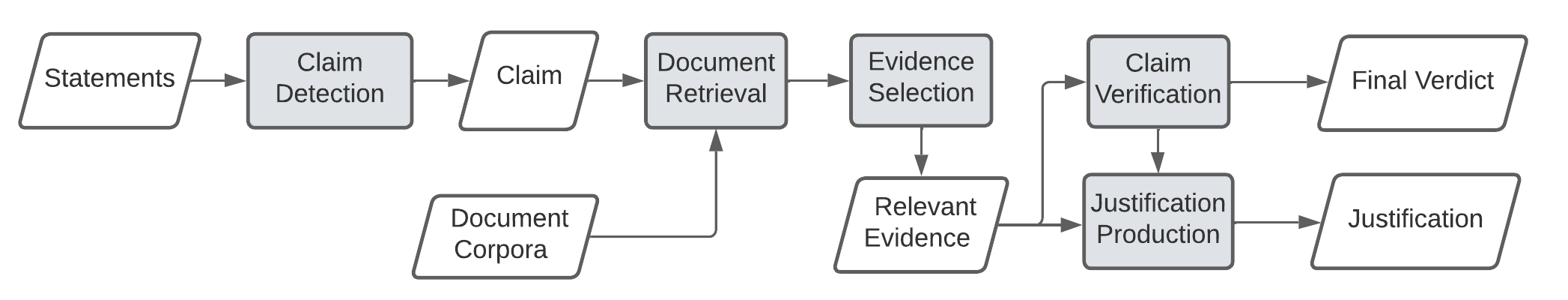}
  \vspace*{-15pt}
  \caption{Overview of a general AFC pipeline. White diamond blocks are documents and objects, and grey square blocks are AFC subtasks. Specific formulations and implementation of course might differ.}
  \label{fig:generic_workflow}
%\vspace*{-15pt}
\end{figure}

%- AFC's general framework
A general framework for AFC has been identified by recent surveys~\cite{zeng2021automated,guo2022survey}, and can be seen in Figure~\ref{fig:generic_workflow}. Zeng et al.~\cite{zeng2021automated} define it as a multi-step process where each step can be tackled as a subtask. Firstly, a \textit{claim detection} step identifies which claims need to be verified. Based on such claims, a \textit{document retrieval} step gathers documents that might contain information relevant to verifying the claim. A \textit{sentence selection} step then identifies and extracts from retrieved documents a set of few individual text passages that are deemed the most relevant. Based on these passages, a \textit{claim verification} step provides the final verdict. Guo et al.~\cite{guo2022survey} add that a final \textit{justification production} step is crucial for explainability. Given the framework's nature, it is no wonder pipelined approaches are extremely popular and compose the current state-of-the-art.

%Figure of the pipeline

%- what normally consist inputs/evidence/outputs
%- how KG factors in (input/evidence) and a segway to AFC on KGs
AFC mainly deals with text, both as claims to be verified and as evidence documents, due to recent advances in this direction being greatly facilitated by textual resources like the FEVER shared task~\cite{thorne2018fact} and its associated large-scale benchmark FEVER dataset~\cite{thorne2018fever}. Still, some tasks in AFC take semantic triples as verifiable claims, either from KGs~\cite{shi2016discriminative,kim2020unsupervised} or by extracting them from text. Some also utilise KGs as reasoning structures from where to draw evidence~\cite{guo2022survey,vlachos2015identification,thorne2017extensible,ciampaglia2015computational,shiralkar2017finding}. For instance, Thorne and Vlachos~\cite{thorne2017extensible} directly map claims found in text to triples in a KG to verify numerical values. Both Ciampaglia et al.~\cite{ciampaglia2015computational} and Shiralkar et al.~\cite{shiralkar2017finding} use entity paths in DBpedia to verify triples extracted from text. Other approaches based on KG embeddings associate the likelihood of a claim being true to that of it belonging as a new triple to a KG~\cite{ammar2019fact,joshi2022ensemble}.

These tasks, while incorporating semantic triples and KGs, can not be exactly defined as AFC on KG; either the verified triples do not come from full and consistent KGs, or the evidence used for reasoning is not taken from sources that could serve as provenance, but inferred from the graph itself.

\subsubsection{AFC on KGs}

AFC on KGs is a more specific task within AFC, explored by a handful of approaches, the most prominent of which are DeFacto~\cite{gerber2015defacto} and its successor FactCheck~\cite{syed2018factcheck}. Its main purpose is to ensure KGs are fit for use by asserting whether their information is verifiable by trustworthy evidence. Given a KG triple and either its documented external provenance or searchable external document corpora whose items could be used as provenance, AFC on KGs can be defined as the automated verification of said triple's veracity or support by collecting and reasoning over evidence extracted from such actual or potential provenance. Its outputs are the verdict and the evidence used.

KGCleaner~\cite{padia2018kgcleaner} uses string matching and manual mappings to retrieve sentences relevant to a KG triple from a document, using embeddings and handcrafted features to predict the triple's credibility. Leopard~\cite{speck2019leopard} validates KG triples for three specific organisation properties, using specifically designed extractions from HTML content. Both approaches entail manual work overhead, cover a limited amount of predicates, and do provide human-readable evidence.

DeFacto~\cite{gerber2015defacto} and its successor FactCheck~\cite{syed2018factcheck} represent the current state-of-the-art on this task. They verbalise KG triples using text patterns and use it to retrieve web pages with related content. They then score sentences based on relevance to the claim and use a supervised classifier to classify the entire web page. Despite their good performance, both approaches depend on string matching, which might miss verbalisations that are more nuanced and also entail considerable overhead for unseen predicates. ProVe, on the other hand, covers any non-ontological predicate (such as \textit{subclass of} and \textit{main category of}) by using pre-trained LMs that leverage context and meaning to infer verbalisations.

Due to its specific application scenario, approaches tackling AFC on KGs differ from the general framework~\cite{zeng2021automated,guo2022survey} seen in Figure~\ref{fig:generic_workflow}. A \textit{claim detection} step is not deemed necessary, as triples are trivial to extract and it is commonly assumed they all need verifying. Alternatively, triples with particular predicates can be easily selected. The existence of a \textit{document retrieval} step depends on whether provenance exists or needs to be searched from a repository, with the former scenario dismissing the need for the step. This is the case for ProVe, but not for DeFacto~\cite{gerber2015defacto} and FactCheck~\cite{syed2018factcheck}, which search for web documents.

Additionally, KG triples are often not understood by the components' main labels alone. Descriptions, alternative labels, editor conventions, and discussion boards help define their proper usage and interpretation, rendering their meaning not trivial, in contrast to the natural language sentences tackled by general AFC. As such, approaches tackling AFC on KGs rely on transforming KG triples into natural sentences~\cite{lehmann2012defacto,gerber2015defacto,syed2018factcheck} through an additional \textit{claim verbalisation} step. While both DeFacto~\cite{gerber2015defacto} and FactCheck~\cite{syed2018factcheck} rely on sentence patterns that are completed with the components' labels, ProVe relies on state-of-the-art Language Models (LMs) for data-to-text conversion.

Lastly, evidence document corpora normally used in general AFC tend to have a standard structure or come from a specific source. Both FEVER~\cite{thorne2018fever} and VitaminC~\cite{schuster2021vitaminc} take their evidence sets from Wikipedia, with FEVER's even coming pre-segmented as individual and clean sentences. Vo and Lee~\cite{vo2020facts} use web articles from \url{snopes.com} and \url{politifact.com} only. KGs, however, accept provenance from potentially any website domains. As such, unlike general AFC approaches, ProVe employs a \textit{text extraction} step in order to retrieve and segment text from triples' references. While previous approaches simply remove HTML markup, ProVe employs a rule-based approach that allows for more flexibility.

\subsubsection{Large Pre-trained Language Models on AFC}

Advances towards textual evidence-based AFC, particularly the \textit{sentence selection} and \textit{claim verification} subtasks, have been facilitated by resources like the FEVER~\cite{thorne2018fact,thorne2018fever} shared task and its benchmarking dataset. The FEVER dataset consists of a large set of claims annotated with one of three classes: supported, refuted, and not enough information to determine (neither). The dataset also provides pre-extracted and segmented passages from Wikipedia as evidence for each claim.

Tackling FEVER through pre-trained LMs~\cite{soleimani2020bert,lewis2020rag,malon2019team} and graph networks~\cite{liu2019kgat,zhou2019gear,zhong2019dream} represents the current state-of-the-art. While approaches using graph networks (such as KGAT~\cite{liu2019kgat}, GEAR~\cite{zhou2019gear}, and DREAM~\cite{zhong2019dream}) for \textit{claim verification} slightly outperform those based mainly on sequence-to-sequence LMs, they still massively depend on the latter for \textit{sentence selection}. Additionally, explainability for task-specific graph architectures, like those of KGAT and DREAM, is harder to tackle than for generalist sequence-to-sequence LM architectures which are shared across the research community~\cite{bracsoveanu2022visualizing,rastas2022explainable,kumar2020explainable}. Slightly decreasing potential performance in favour of a simpler and more explainable pipeline, ProVe employs LMs for both \textit{sentence selection} and \textit{claim verification}.

On \textit{sentence selection}, the common strategy is to assign relevance scores to text passages based on their contextual proximity to a verifiable claim. GEAR~\cite{zhou2019gear} does so with LSTMs, but use a BERT model to acquire text encodings. Soleimani et al.~\cite{soleimani2020bert}, KGAT~\cite{liu2019kgat}, and current state-of-the-art DREAM~\cite{zhong2019dream} outperform GEAR by directly using BERT for the rankings, an approach ProVe also follows. Graph networks are employed at the \textit{claim verification} subtask~\cite{liu2019kgat,zhou2019gear,zhong2019dream}. Soleimani et al.~\cite{soleimani2020bert} are among the few to achieve near state-of-the-art results using an LM and a rule-based aggregation instead of graph networks. While ProVe handles the subtask similarly, it uses a weak classifier as its final aggregation.

As a subtask of AFC on KGs, \textit{claim verbalisation} is normally done through text patterns~\cite{gerber2015defacto,syed2018factcheck} and by filling templates~\cite{padia2018kgcleaner}, both of which can either be distantly learned or manually crafted. ProVe is the first approach to utilise an LM for this subtask. Amaral et al.~\cite{amaral2022wdv} shows that a T5 model fine-tuned on WebNLG achieves very good results when verbalising triples from Wikidata across many domains. ProVe follows suit by also using a T5.

%\comment{
Table~\ref{tab:approaches} shows a comparison of ProVe to other AFC approaches mentioned in this section grouped by specific task, showcasing the particular subtasks each targets, as well as the datasets used as a basis for their evaluation. AFC on KG is amongst the least researched tasks within AFC. ProVe is the first to tackle it through fine-tuned LMs that adapt to unseen KG predicates and to be evaluated on a Wikidata dataset consisting of multiple non-ontological predicates.

% This table is very nice. Perhaps add a column with the methods used and the main things you learned from those papers or how they relate to your work?

\begin{table}[ht]
\centering
\begin{tabular}{|l|l|l|l|l|l|l|}
\hline
Task &
  \begin{tabular}[c]{@{}l@{}}Input\\ Type\end{tabular} &
  \begin{tabular}[c]{@{}l@{}}Evidence\\ Source\end{tabular} &
  \begin{tabular}[c]{@{}l@{}}Evidence\\ Returned\end{tabular} &
  Subtasks &
  \begin{tabular}[c]{@{}l@{}}Evaluation\\ Dataset\end{tabular} &
  Approaches \\ \hline
\begin{tabular}[c]{@{}l@{}}General\\ text-based\\ AFC\end{tabular} &
  \begin{tabular}[c]{@{}l@{}}Textual\\ claims\end{tabular} &
  Text &
  Yes &
  \begin{tabular}[c]{@{}l@{}}DR,\\ SS,\\ CV\end{tabular} &
  FEVER &
  \begin{tabular}[c]{@{}l@{}}\cite{soleimani2020bert,malon2019team,liu2019kgat,zhou2019gear,zhong2019dream},\end{tabular} \\ \hline
\begin{tabular}[c]{@{}l@{}}Graph-based\\ AFC\end{tabular} &
  \begin{tabular}[c]{@{}l@{}}Textual\\ claims\end{tabular} &
  KG &
  Yes &
  \begin{tabular}[c]{@{}l@{}}SS,\\ CV\end{tabular} &
  Freebase &
  \begin{tabular}[c]{@{}l@{}}\cite{vlachos2015identification,thorne2017extensible}\end{tabular} \\ \hline
\multirow{2}{*}{\begin{tabular}[c]{@{}l@{}}KG triple\\ prediction\end{tabular}} &
  \multirow{2}{*}{\begin{tabular}[c]{@{}l@{}}KG\\ triples\end{tabular}} &
  \begin{tabular}[c]{@{}l@{}}KG\\ paths\end{tabular} &
  Yes &
  \begin{tabular}[c]{@{}l@{}}SS,\\ CV\end{tabular} &
  \begin{tabular}[c]{@{}l@{}}DBpedia,\\ SemMedDB,\\ Wikipedia\end{tabular} &
  \begin{tabular}[c]{@{}l@{}}\cite{shi2016discriminative,kim2020unsupervised,ciampaglia2015computational,shiralkar2017finding},\end{tabular} \\ \cline{3-7} 
 &
   &
  KGE &
  No &
  \begin{tabular}[c]{@{}l@{}}RE,\\ EL,\\ CV\end{tabular} &
  \begin{tabular}[c]{@{}l@{}}Kaggle,\\ news articles\\ DBpedia,\\ Freebase\end{tabular} &
  \begin{tabular}[c]{@{}l@{}}\cite{ammar2019fact,joshi2022ensemble}\end{tabular} \\ \hline
\multirow{3}{*}{\begin{tabular}[c]{@{}l@{}}AFC\\ on KGs\end{tabular}} &
  \multirow{3}{*}{\begin{tabular}[c]{@{}l@{}}KG\\ triples\end{tabular}} &
  \multirow{3}{*}{Text} &
  Yes &
  \begin{tabular}[c]{@{}l@{}}CVb,\\ DR,\\ TA,\\ CV\end{tabular} &
  \begin{tabular}[c]{@{}l@{}}DBpedia,\\ FactBench\end{tabular} &
  \begin{tabular}[c]{@{}l@{}}\cite{lehmann2012defacto,gerber2015defacto,syed2018factcheck}\end{tabular} \\ \cline{4-7} 
 &
   &
   &
  No &
  \begin{tabular}[c]{@{}l@{}}SS,\\ CV\end{tabular} &
  \begin{tabular}[c]{@{}l@{}}Wikidata\\ (48 predicates),\\ SWC 2017\end{tabular} &
  \begin{tabular}[c]{@{}l@{}}\cite{padia2018kgcleaner,speck2019leopard}\end{tabular} \\ \cline{4-7} 
 &
   &
   &
  Yes &
  \begin{tabular}[c]{@{}l@{}}CVb,\\ TR,\\ SS,\\ CV\end{tabular} &
  \begin{tabular}[c]{@{}l@{}}Wikidata\\ (any non-ontological\\ predicate)\end{tabular} &
  ProVe \\ \hline
\end{tabular}
\vspace{5pt}
\caption{Comparison between ProVe and others within AFC. KGE = KG Embeddings, DR = Document Retrieval, SS = Sentence Selection, CV = Claim Verification, RE = Relation Extraction, EL = Embedding Learning, CVb = Claim Verbalisation, TA = Trustworthiness Analysis, TR = Text Retrieval.}
%\vspace{-20pt}
\label{tab:approaches}
\end{table}

%
%\newpage
\section{Approach}
\label{sec:method}
%Methodology (had 7.5, now 9)

%You need a formalisation of what you did, of the task, of the metrics, of the components in Section 3 etc. Use numbers in Figure 3 that match to the headings 3.2, 3.3, 3.4, 3.5. This is a good section, but it lacks formulas..

%Formalisation of:
%- Task
%- Metrics
%- Components
    %- Formulas
%Add numbers to figure 3 pointing to the subsection headings
%Assure that subsubsections will not be needed

ProVe consists of a pipeline for Automated Fact Checking (AFC) on Knowledge Graphs (KG) that, provided with a KG triple that is not ontological in nature (e.g. denoting subclasses, categories, lists, etc) and its documented provenance in the form of a web page or text document, automatically verifies whether the page textually supports the triple, retrieving from it relevant text passages that can be used as evidence. This section presents an overview of ProVe and its task, as well as detailed descriptions of its modules and the subtasks they target.

\subsection{Overview}

From a KG's set of non-ontological triples $T$, consider a KG triple $t \in T$, where $t$ is composed by a subject, a predicate, and an object, i.e. $t = (s,p,o)$. Consider also a reference $r$ to a web page or text document, acting as the documented provenance of $t$. ProVe assesses whether $r$ textually supports $t$ through a pipeline of rule-based methods and language models. 

\begin{figure}[h]
  \centering
  %\vspace*{-15pt}
  \includegraphics[width=1\linewidth]{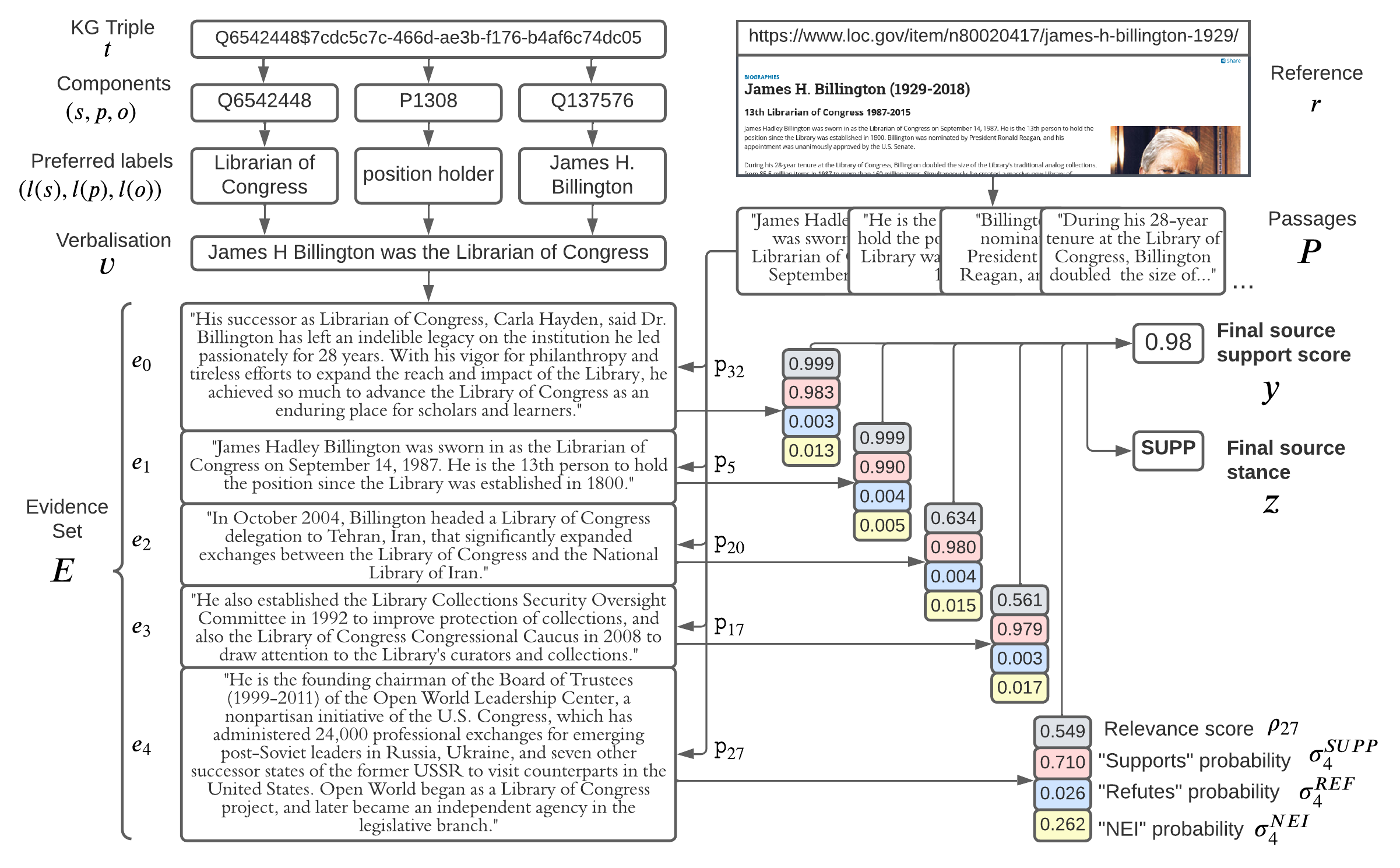}
  %\vspace*{-15pt}
  \caption{An example of the inputs and outputs of ProVe when applied to a Wikidata triple and its provenance. A triple's ($t$) subject, predicate, and object elements ($s,p,o$) have their labels extracted and verbalised ($v$). Reference $r$ has its passages extracted ($P$). The 5 most relevant passages are compiled as the evidence set ($E$), with their respective relevance scores ($\rho_{i}$) and stance probabilities ($\sigma_{i}^k$) used to calculate a final class ($z$) and a support probability ($y$). Note that $i$ indices between passages in $P$ and evidence in $E$ are different, i.e. the $27^{th}$ extracted passage is the $4^{th}$ evidence.}
  \label{fig:example}
%\vspace*{-15pt}
\end{figure}

Figure~\ref{fig:example} shows a KG triple (taken from Wikidata), its reference to a web page, and ProVe's processing according to the definitions provided in this section. ProVe extracts the text from $r$ and divides it into a set of passages $P$. Each passage $\mathtt{p}_{i} \in P$, with $i$ in the closed integer interval $[0..|P|-1]$, receives a relevance score $\rho_{i} \in [-1,1]$ indicating how relevant they are to the triple $t$. The five highest-ranking passages from $P$ are selected as evidence and assembled into the evidence set $E$. Each evidence $e_{i} \in E$ receives three stance probabilities $(\sigma^{SUPP}_{i}, \sigma^{REF}_{i}, \sigma^{NEI}_{i})$, with $i$ in the integer interval $[0..4]$, such that each probability $\sigma^{k}_{i} \in [0,1]$ and $\sum \sigma^{k}_{i} = 1$ for $k \in \mathcal{K}$ and $\mathcal{K} = \{SUPP, REF, NEI\}$. These probabilities denote the individual stance of evidence $e_{i}$ towards the triple $t$, which can either support the triple ($SUPP$), refute it ($REF$), or not have enough information to do either ($NEI$). Finally, ProVe uses the relevance scores $\rho_{i}$ and the stance probabilities $\sigma^{k}_{i}\ \forall\ k \in \mathcal{K}$ of each $e_i \in E$ to define a final stance $z \in \mathcal{K}$, as well as to calculate a support probability $y \in [0,1]$ indicating how much the triple $t$ is supported by its reference $r$. 

\begin{figure}[ht]
  \centering
  %\vspace*{-15pt}
  \includegraphics[width=1\linewidth]{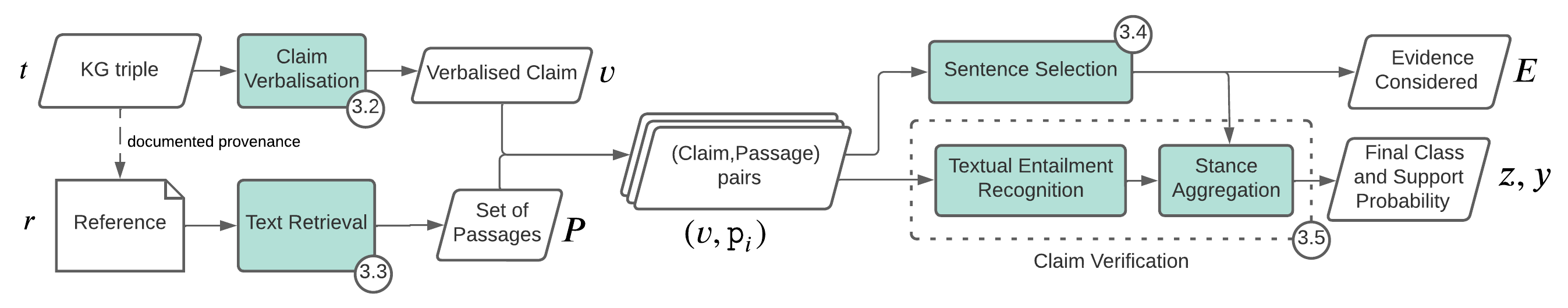}
  %\vspace*{-15pt}
  \caption{Overview of ProVe's pipeline. The white blocks are artefacts while the green are modules, further detailed in the subsections indicated in the circles.}
  \label{fig:workflow}
%\vspace*{-15pt}
\end{figure}

A modular view of ProVe's pipeline can be seen in Figure~\ref{fig:workflow}. Its inputs, as previously stated, are a KG triple ($t$) and a referenced web page or text document ($r$), while its outputs are a final stance class ($z$), a support probability ($y$), and a set of textual evidence used to calculate it ($E$). ProVe takes any non-ontological KG triple as long as its components are accompanied by labels in natural language. The claim verbalisation module takes the preferred labels of each of the triple's components (i.e. subject, predicate, and object) as its inputs and produces a natural language sentence that expresses the same information ($v$). As KG entities and predicates might contain multiple labels, multiple possible verbalisations can be generated; users might chose those labels that best portray the triple's meaning, which are here defined as preferred labels, and even rules within the KG can determine them. The reference ($r$) can consist of any HTML page containing natural language text, but also plain text documents. As ProVe makes no assumptions as to the page's layout in order to optimise recall, its text retrieval module extracts all identified passages ($P$) from the page, even if they contain boilerplate text or have poor syntax due to layout-dependant contents, e.g. tables, headers, etc. 

Pairs consisting of the verbalised claim ($v$) alongside each extracted passage ($\mathtt{p}_{i}$) are given to the sentence selection and the claim verification modules. At the former, extracted passages are given a relevance score from -1 to 1, indicating how contextually relevant to the claim they are, regardless of stance. The five highest scoring passages are selected as evidence ($E$). The claim verification module has two steps. First, a Textual Entailment Recognition (TER) step assigns each extracted passage with probabilities of having the three following stances: supports the claim, refutes the claim, or does not hold enough information for either (NEI). Finally, a stance aggregation step takes the relevance scores plus stance probabilities for the five passages in the evidence set and outputs the final class ($z$) and support probability ($y$), indicating how supportive of the triple is the reference.

ProVe's workflow differs from the AFC framework seen in Figure~\ref{fig:generic_workflow}. This is due to the particular task ProVe tackles, i.e. the verification of KG triples using text from documented provenance, where triples can have any non-ontological predicate and such provenance can come from varied sources. As detailed in Section~\ref{sec:related} and evidenced in Table~\ref{tab:approaches}, this is currently a little studied task compared to others in AFC, posing distinct problems and requiring specific subtasks to be solved. For instance, ProVe does not need to perform either claim detection or document retrieval, as both the claims and the sources are given to it as inputs, although such modules can be easily plugged in from other pipelines.

On the other hand, as ProVe handles both KG triples and unstructured text with the same model architectures and does not make use of KG paths for evidence, it needs to convert the triples into text through a claim verbalisation module, akin to most other approaches in this task~\cite{lehmann2012defacto,gerber2015defacto,syed2018factcheck,padia2018kgcleaner}. As its input references can lead to web pages having any HTML layout and their text does not come pre-segmented (such as with FEVER~\cite{thorne2018fever}), ProVe needs a non-trivial text retrieval module so that it can identify informative passages. Finally, as such KGs are often secondary sources of information and triples should not include conclusions or interpretations from editors, ProVe needs to consider pieces of evidence in isolation; this is to lower ProVe's reliance on multi-sentence reasoning, as concluding a triple from multiple text passages should not configure support in this task. Thus, ProVe first identifies stances of retrieved evidence individually, aggregating them into a final verdict afterwards. Each of ProVe's modules is further detailed in the remainder of this section.

\subsection{Claim Verbalisation}
\label{sec:method_verb}

KG entities and properties have natural language labels that help clarify their meanings, with KGs like Wikidata and Yago also containing multiple aliases and alternative names. However, these entities and predicates are often not created by prioritising human understanding, but rather data organisation, and thus rely heavily on descriptions in order to set out proper usage rules. Many serve as abstract concepts that unite other related but not identical concepts, using a very broad main label and more specific aliases. One example of such is Wikidata's \textit{inception} property (\textit{P571}), which indicates the date in which something was founded, written, formed, painted, or created, and applies to any entity with a beginning; its description clearly points out that for dates of official opening, \textit{P1619} should be used instead. Many also depend heavily on context (e.g. subject and object types) or editor conventions to have a clear meaning. One example of such is the \textit{child} property (\textit{P40}), which follows the convention that the subject has the object as its child, but should not be used for stepchildren. However, the triple \textit{$($John,\ child,\ Paul$)$} alone makes it unclear which is the parent. As such, merely concatenating labels does not convey the full meaning of the triple~\cite{gerber2015defacto,lehmann2012defacto,syed2018factcheck}. Thus, ProVe relies on a claim verbalisation component that, based on similar triples and their verbalisations, is able to fill format and meaning gaps.

ProVe's claim verbalisation module takes as input a KG triple $t$, made by its subject, predicate, and object components such that $t = (s, p, o)$. A component's preferred label is assumed to be its main (and often only) label, but alternative labels or aliases can be manually chosen by editors and curators employing ProVe. Denoting by the function $l(\cdot)$ the process of defining a component's preferred label, ProVe's claim verbalisation module outputs a natural language formulation $v$, called the verbalisation, as defined by Equation \ref{eq:claim_verb}.

\begin{equation}
v = \phi(l(s), l(p), l(o))
\label{eq:claim_verb}
\end{equation}

The function $\phi(\cdot)$ represents the generation of a fluent and adequate verbalisation from the components' preferred labels. A fluent verbalisation is defined as one written in good grammar, resembling natural text, and an adequate verbalisation as one that carries the same meaning as the original triple intended. Like recent works in data verbalisation~\cite{ribeiro2020investigating,amaral2022wdv}, a pre-trained transformer is used for this subtask. The function $\phi(\cdot)$ is carried out by a T5-base~\cite{raffel2020exploring} model fine-tuned on the WebNLG 2017 dataset~\cite{gardent2017webnlg}. The WebNLG 2017 dataset consists of DBpedia triples belonging to 15 categories and their corresponding verbalisations in English; 10 categories (the \textit{SEEN} partition) were used for training, validation, and testing, and the remaining 5 (the \textit{UNSEEN} partition) for testing only. Fine-tuning was carried out with a constant $3\text{e-}5$ learning rate, an Adam optimiser, and a cross-entropy loss for 100 epochs with early stopping patience of 15 epochs. Its text generation is done via beam search with 3 beams.

Amaral et al.~\cite{amaral2022wdv} use this exact same model to produce the WDV dataset, which consists of verbalised Wikidata triples, and then evaluate its quality with human annotations on both fluency and adequacy. Such evaluations are covered in more detail in Section~\ref{sec:eval}. In WDV, main labels are used as preferred labels for all triple components, despite aliases often representing better choices. ProVe allows editors to manually define the behaviour of $l(\cdot)$ to replace contextually-dependent and vague main labels with alternate labels in order to address some of the fluency and adequacy issues observed in WDV. While this entails the extra effort by ProVe's users of choosing proper labels, it is still a much more scalable alternative to manually generating verbalisations.

\subsection{Text Retrieval}
\label{sec:method_text}

In KGs, provenance is documented per individual triples and is often presented as URLs to web pages or text documents. Such references form the basis of KG verifiability and should point to sources of evidence that support their associated KG triples. Additionally, they can come from a huge variety of domains as long as they adhere to verifiability criteria, that is, humans can understand the information they contain. As humans are excellent in making sense of structured and unstructured data combined, KG editors do not need to worry much about how references express their information. Images, charts, tables, headers, infoboxes, and unstructured text, can all serve as evidence to the information contained in KG triple. However, this complicates the automated extraction of such evidence in a standard format so that LMs can understand it.

\begin{figure}[ht]
  \centering
  %\vspace*{-15pt}
  \includegraphics[width=1\linewidth]{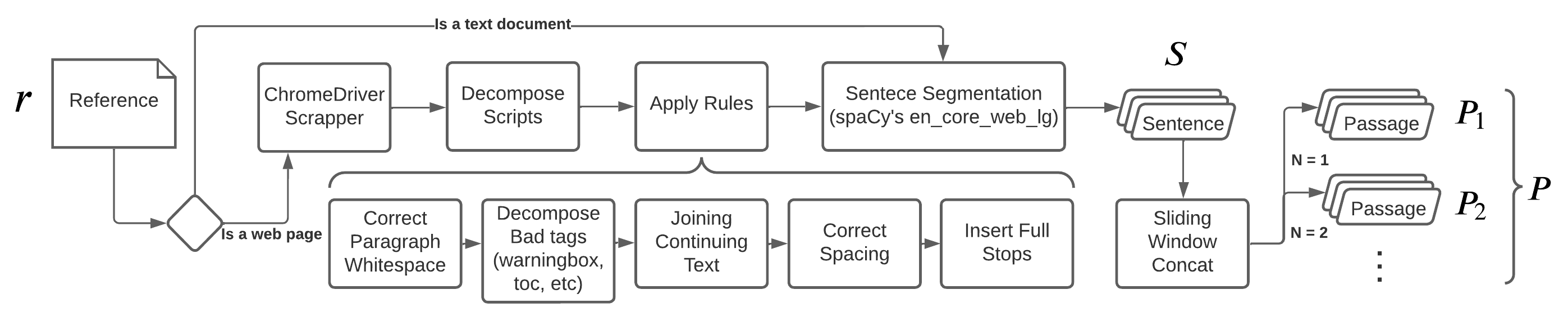}
  %\vspace*{-15pt}
  \caption{Illustration of the text extraction module's workflow, taking a reference $r$ as input, dividing its text into a set of segments $S$, and concatenating them into a set of extracted passages $P$.}
  \label{fig:text_extractor}
%\vspace*{-15pt}
\end{figure}

Rather than only free-flowing text, referenced web pages can have multiple sections, layouts, and elements, making it non-trivial to automatically segment its textual contents into passages. Thus, ProVe employs a combination of rule-based methods and pre-trained sentence segmenters to extract passages. Figure~\ref{fig:text_extractor} details this process. The module takes as input a reference $r$, which can either be a URL to a web page or a text document. If an URL, the module extracts all HTML contents via a web scrapper, assuring that all contents accessible by users are rendered and processed. The module then removes scripts and code from the HTML, leaving only markup and text. A list of rule-based cleaning steps are then applied. Whitespaces in continuous text elements are corrected by ensuring text within tags such as \textit{$<$p$>$} do not have separations that could be breaking sentences. Tags that are sure to be boilerplate are removed, such as tables of contents and navigation bars. Text that is broken across sequential similar tags is joined. Lastly, spacing and punctuation (full stops) are corrected. Following this process addresses the most severe cases of improper sentence segmentation. Leftover HTML markup is removed and the text text is fed into spaCy's sentence segmenter using the $en\_core\_web\_lg$ model~\cite{spacy}, producing a set of text segments $S$. If $r$ is a text document, it is fed directly into sentence segmentation.

As a last step, multiple $n$-sized sliding window concatenations are used to create the final set of passages $P$. Given a positive integer $n \in \mathcal{N}$, where $\mathcal{N}$ is the set of window sizes to be applied, an $n$-sized window slides through the sequential set $S$ of text segments produced by the sentence segmentation step. 
Let $S^i_j=[s_i,\ldots,s_j]$, for $j\geq i$, be the window including all segments from $s_i$ to $s_j$ and 
$\odot(\cdot)$ the function that concatenates a sequence of segments interleaving them with a blank space as a separator. 
% Considering the window starts at the $i$-th segment $\mathtt{s}_{i} \in S$, it concatenates all $n$ segments in the window ($\mathtt{s}_{i}$ to $\mathtt{s}_{i+n-1}$), with a space as separator, into a passage, with $j$ in the integer interval $[0..|S|-n]$.
Equations~\ref{eq:slide} and ~\ref{eq:pass} define the set of all passages $P_n$ produced by a $n$-sized sliding window and the union $P$ of all such passages, respectively.

% \begin{equation}
% \begin{gathered}
% \mathtt{p}^n_{i} = \odot(\mathtt{s}_{i},...,\mathtt{s}_{i+n-1})
% \label{eq:slide}
% \end{gathered}
% \end{equation}

\begin{equation}
\begin{gathered}
P_n = \{\odot(S^i_{i+n-1})\; |\; 0 \leq i \leq |S| -n\}
\label{eq:slide}
\end{gathered}
\end{equation}

\begin{equation}
\begin{gathered}
P = \bigcup_{n \in \mathcal{N}} P_n
\label{eq:pass}
\end{gathered}
\end{equation}

ProVe concatenates text segments in this fashion for two reasons. Firstly, meaning can often be spread between sequential segments, e.g. in the case of anaphora. Secondly, HTML layouts might separate text in ways ProVe's general rules were not able to join, e.g. a paragraph describing an entity in the header. For a trade-off between coverage and sentence length, ProVe defines $\mathcal{N} = \{1,2\}$, i.e. it concatenate all segments by themselves ($P_1$) and all sequential pairs ($P_2$).

Current best approaches to extracting textual content from web pages, often called boilerplate removal, are based on supervised classification~\cite{vogels2018web2text,leonhardt2020boilerplate} and, as no classification is perfect, might miss relevant text. ProVe's text retrieval module aims at maximizing recall by retrieving all reachable text from the web page and arranging it into separate passages by following a set of rules based on the HTML structure. The sentence selection module is later responsible for performing relevance-based filtering. ProVe's rule-based method can easily be updated with ad hoc cleaning techniques to help treat difficult cases, such as linearisation or summarisation of tables, automated image and chart descriptions, converting hierarchical header-paragraph structures and infoboxes into linear text, etc. 

\subsection{Sentence Selection}
\label{sec:method_sensec}

As ProVe's text extraction module extracts all the text in a web page in the form of passages $P$, it needs a way to filter those based on their relevance to the triple $t$. Sentence selection consists of, given a claim, rank a set of sentences based on how relevant each is to the claim, where relevance is defined as contextual proximity e.g. similar entities or overlapping information. Sentence selection is an integral part of most recent AFC approaches~\cite{guo2022survey,zeng2021automated}, incuding AFC on KGs~\cite{syed2018factcheck,gerber2015defacto}, as discussed in Section~\ref{sec:related_afc}. It is a subtask of the FEVER fact-checking shared task~\cite{thorne2018fact}, directly supported as a supervised task by the FEVER dataset~\cite{thorne2018fever}, and is explored by a large body of work~\cite{soleimani2020bert,malon2019team,liu2019kgat,zhou2019gear,zhong2019dream} to excellent results.

Following on KGAT's~\cite{liu2019kgat} and DREAM's~\cite{zhong2019dream} approach to FEVER's sentence selection subtask, ProVe employs a large pre-trained BERT transformer. ProVe's sentence selection BERT is fine-tuned on the FEVER dataset by adding to it a dropout and a linear layer, as well as a final hyperbolic tangent activation, making outputted scores range from $-1$ to $1$, and training it on a pairwise margin ranking loss with the margin set to $1$. Fine-tuning is achieved by feeding the model pairs of inputs, where the first element is a concatenation of a claim and a relevant sentence, while the second element is the same but with an irrelevant sentence instead, and training it to assign higher scores to the first element, such that the difference in scores between the pair is $1$ (the margin). FEVER is used for training and validation. For each FEVER claim in the training and validation partition, relevant sentences are provided as annotations. Irrelevant sentences were retrieved by applying the same document retrieval process used by other works~\cite{soleimani2020bert,liu2019kgat,zhou2019gear} to define relevant Wikipedia articles, which FEVER breaks into pre-segmented sentences. All sentences from such retrieved documents that were not already annotated as relevant were taken as irrelevant.

This fine-tuned module is thus used to assign scores ranging from $-1$ to $1$ to passages from $P$, expressing how relevant they are to the given triple $t$. Taking the triple $t$'s verbalisation $v$ as the claim, and a passage $\mathtt{p}_i$ as the sentence, the BERT model takes the concatenation of $v$ and $\mathtt{p}_i$ as input and outputs a relevance score $\rho_i \in [-1,1]$. This is defined in Equation~\ref{eq:sent}, where $\psi(\cdot)$ represents the execution of the sentence selection BERT on the concatenated input. 

ProVe ranks all passages $\mathtt{p}_i \in P$ based on their relevance scores $\rho_i$. As passages are generated with sliding windows of different sizes by the text extraction module, they might have overlapping content. To avoid unnecessary repetition of information, a passage $\mathtt{p}_i$ is removed from $P$ whenever there is another passage $\mathtt{p}_j$ that overlaps with it and is more relevant (i.e, $\rho_j > \rho_i$), yielding the set of passages $P^*$, whose five highest scoring passages constitute the evidence set $E$ (see Equation~\ref{eq:evidence}).

\begin{equation}
\begin{gathered}
\rho_i = \psi(v, \mathtt{p}_i) 
\label{eq:sent}
\end{gathered}
\end{equation}

\begin{equation}
\begin{gathered}
E = \argmax_{P' \subseteq P^{*}, |P'|=5} \sum_{\mathtt{p}_i \in P'} \rho_i
\label{eq:evidence}
\end{gathered}
\end{equation}

\subsection{Claim Verification}
\label{sec:method_claimver}

As discussed in Section~\ref{sec:related_afc}, claim verification is a crucial subtask in AFC, central to various approaches~\cite{guo2022survey,thorne2018automated}, and consists on assigning a final verdict to a claim, be it on its veracity or support, given retrieved relevant evidence. ProVe's claim verification relies on two steps: first, a pre-trained BERT fine-tuned on data from FEVER~\cite{thorne2018fever} performs Textual Entailment Recognition (TER) to detect stances of individual pieces of evidence, and then an aggregation considers the stances and relevance scores from all evidence to define a final verdict. As ProVe also uses a BERT model for sentence selection, its approach is similar to that of Soleimani et al.~\cite{soleimani2020bert}, which uses two fine-tuned BERT models, one for sentence selection, another for claim verification. Although task-specific graph-based approaches~\cite{liu2019kgat,zhong2019dream} outperform Soleimani, it is by less than a percentage point (on FEVER score), while explainability for such generalist pre-trained LMs is increasingly researched~\cite{bracsoveanu2022visualizing,rastas2022explainable,kumar2020explainable} by the NLP community.

\subsubsection{Textual Entailment Recognition}

Like sentence selection, claim verification is a well defined subtask of AFC, supported by both the FEVER shared task and the FEVER dataset. FEVER annotates claims as belonging to one of three TER classes: those supported by their associated evidence (`SUPPORTS'), those refuted by it (`REFUTES'), and those wherein the evidence is not enough to reach a conclusion (`NOT ENOUGH INFO', abbreviated as `NEI'). As previously mentioned, ProVe is meant to handle KGs as secondary sources of information. Thus, it assesses evidence first in isolation, aggregating assessments afterwards in a similar fashion to other works~\cite{soleimani2020bert,malon2019team}.

ProVe's TER step is a BERT model fine-tuned on a multiclass classification TER task. It consists of identifying a piece of evidence's stance towards a claim using the three classes from FEVER (`SUPPORTS',`REFUTES',`NEI'). To fine-tune such model, a labelled training dataset of claims-evidence pairs is built out of FEVER. For each claim in FEVER labeled as `SUPPORTS', all sentences annotated as relevant to it are paired with the claim; such pairs are labelled as `SUPPORTS'. The same is done for all claims in FEVER labeled as `REFUTES', generating pairs classified as `REFUTES'. For claims labeled as `NOT ENOUGH INFO', FEVER does not annotate any sentence as relevant to them. Thus, ProVe's sentence selection module is applied to documents deemed relevant to such claims (retrieved in a similar fashion to KGAT~\cite{liu2019kgat}) and each claim is paired with all sentences that have relevance scores greater than $0$ in regards to them. All such pairings are labelled `NEI'. Fine-tuning was carried for $2$ epochs with an AdamW optimizer with $0.01$ weight decay. Population Based Training was used to tune learning rate, batch sizes, and warmup ratio.

Thus, for a verbalisation $v$ obtained from a triple $t$ and a piece of evidence $e_i \in E$, retrieved as a passage from the $t$'s reference $r$, ProVe's TER step returns a probability array $\sigma_{i}$ that describes $e_i$'s stance towards $v$. Given that $v$ is fluent and adequate, $\sigma_{i}$ also describes $e_i$'s stance towards $t$. Equation~\ref{eq:ter} formulates this, where $\tau(\cdot)$ is the function representing ProVe's TER BERT model, which takes the concatenation of $v$ and $e_i$ as input and outputs $\sigma_{i}$, an array that is normalised through a softmax layer. The array $\sigma_{i}$ consists of the probabilities of each FEVER class $k \in {\cal K}=\{SUPP, REF, NEI\}$. Notice that $\sum_{k \in \mathcal{K}} \sigma_{i}^{k} = 1$.
\begin{equation}
\begin{gathered}
\sigma_{i} = (\sigma_{i}^{SUPP},\sigma_{i}^{REF},\sigma_{i}^{NEI}) = \tau(e_i, v)
\label{eq:ter}
\end{gathered}
\end{equation}

\subsubsection{Stance Aggregation}

After classifying the stance and relevance of each individual piece of evidence $e_i \in E$ towards the triple $t$, ProVe aggregates these scores ($\rho_i$ and $\sigma_i$) into a final stance class $z \in \mathcal{K}$ and a probability $y$ denoting the level of support shown by the triple-reference pair $(t,r)$. Multiple aggregation strategies can be adopted, with ProVe proposing three:

%work here
\begin{enumerate}
    \item A simple weighted sum $\sigma$ of the TER probability arrays $\sigma_i$ using the relevance scores $\rho_i$ as weights, where negative relevance scores are dismissed (Equation~\ref{eq:wsum}). A final TER class $z$ can be defined as the class $k \in \mathcal{K}$ with the highest weighted sum (Equation~\ref{eq:wsum2}). The support probability $y$ is then defined as the weighted summed probability of the `SUPPORTS' class (Equation~\ref{eq:wsum3}). 
    
    \begin{equation}
    \begin{gathered}
    \sigma^k = \sum_{0 \leq i \leq |E|} max(\rho_i,0) * \sigma^k_i% \mid k \in \mathcal{K}
    \label{eq:wsum}
    \end{gathered}
    \end{equation}
    
    \begin{equation}
    \begin{gathered}
    z = \argmax_{k \in \mathcal{K}} \sigma^k
    \label{eq:wsum2}
    \end{gathered}
    \end{equation}
    
    \begin{equation}
    \begin{gathered}
    y = \sigma^{SUPP}
    \label{eq:wsum3}
    \end{gathered}
    \end{equation}
    
    \item The rule-based strategy adopted by Malon et al.~\cite{malon2019team} and Soleimani et al.~\cite{soleimani2020bert}. A triple-reference pair $(t,r)$ is assigned a final TER class $z$ (Equation~\ref{eq:malon1}) of `SUPPORTS` if any individual evidence $e_i \in E$ is most likely to support the triple. If that is not the case, $z$ is set as `REFUTES' if any individual evidence $e_i \in E$ is most likely to refute the triple. If that is also not the case, $z$ is set to `NEI'. The final probability $y$ is $1$ is the triple-reference is classified as `SUPPORTS' and $0$ otherwise (Equation~\ref{eq:malon2}). This strategy does not allow editors to vary the classification threshold.
    
    \begin{equation}
    \begin{gathered}
    z = 
    \begin{cases}
    SUPP,   & \text{if there exists~} e_i \in E \text{~s.t.~} \argmax_{k \in \mathcal{K}} \sigma^k_i = SUPP         \\
    REF,    & \text{else if there exists~} e_i \in E \text{~s.t.~} \argmax_{k \in \mathcal{K}} \sigma^k_i = REF       \\
    NEI,    & \text{otherwise}
    \end{cases}
    \label{eq:malon1}
    \end{gathered}
    \end{equation}
    
    \begin{equation}
    \begin{gathered}
    y = 
    \begin{cases}
    1,& \text{if } z = SUPP\\
    0,              & \text{otherwise}
    \end{cases}
    \label{eq:malon2}
    \end{gathered}
    \end{equation}
    
    \item A simple classifier, trained on an annotated set of triple-reference pairs $(t,r)$. The classifier takes as features all relevance scores $\rho_i$ and all TER probability arrays $\sigma_i$ calculated from every piece of evidence $e_i \in E$, as well as their sizes in characters. It is trained on a multiclass classification task to predict the annotated final TER class $z$ of the pair $(t,r)$ by outputting a probability array $\theta$, as defined by Equation~\ref{eq:class1}, where $\omega(\cdot)$ represents the classifier. Thus, $z$ can be defined as the class with the highest probability (Equation~\ref{eq:class2}), and $y$ as the probability $\theta^{SUPP}$ assigned to the `SUPPORTS' class (Equation~\ref{eq:class3}).
    
    \begin{equation}
    \begin{gathered}
    \theta = (\theta^{SUPP},\theta^{REF},\theta^{NEI}) = \omega( \{ (\rho_i, \sigma_i, |e_i|) \mid e_i \in E \} )
    \label{eq:class1}
    \end{gathered}
    \end{equation}
    
    \begin{equation}
    \begin{gathered}
    z = \argmax_{k \in \mathcal{K}} \theta^k
    \label{eq:class2}
    \end{gathered}
    \end{equation}
    
    \begin{equation}
    \begin{gathered}
    y = \theta^{SUPP}
    \label{eq:class3}
    \end{gathered}
    \end{equation}
    
\end{enumerate}
%
%\newpage
\section{Reference Evaluation Dataset}
\label{sec:data}

%Elena: Here is where I would mention what Wikidata is, not in the abstract or introduction.

This section presents and describes the dataset used to evaluate ProVe: Wikidata Textual References (WTR). WTR is mined from Wikidata, a large scale multilingual and collaborative KG, produced by voluntary anonymous editors and bots, and maintained by the Wikimedia Foundation~\cite{Vrandecic2012}. WTR consists of a series of detailed and annotated triple-reference pairs, each consisting of a non-ontological Wikidata triple paired with a reference to a web page, documented as its provenance. Unlike other benchmarking datasets used in AFC, WTR contains no artificial data and reflects only naturally occurring information. WTR's triples are detailed with all three main components (subject, predicate, and object), as well as their unique Wikidata identifiers, main labels, aliases (alternative labels), and textual descriptions. WTR's references are detailed with the URLs they resolve to, as well as the HTML contents within. WTR is balanced in terms of web domains contemplated, meaning the web domains represented by its references have mostly an equal number of triple-reference pairs.

Each triple-reference pairing in WTR is annotated both at evidence-level and at reference-level. Evidence-level annotations are provided by crowd-workers and describe the stance that specific text passages from the reference display towards the triple. Reference-level annotations are provided by the authors and describe the stance the whole referenced web page displays towards the triple. Evaluation, described in Section~\ref{sec:eval}, consists of comparing ProVe's final class ($z$) and support probability ($y$) outputs to such annotations. Section~\ref{sec:data_con} covers WTR's construction, while Section~\ref{sec:data_ann} details its annotation process.

%Dataset creation
\subsection{Dataset Construction}
\label{sec:data_con}

%Elena: Yes, you said that earlier. I would rephrase this somewhat because it suggests that your system only works in this particular case where there is provenance. Actually quite a few KGs have provenance. It has nothing to do with Wikidata being user-generated or collaboratively generated. Most KGs keep track of where the data from each triple comes from, even if the provenance is not always external references, but rather internal ones.

Wikidata has been chosen as the source for ProVe's evaluation dataset, as it contains vast amounts of triples that explicitly state their provenance, pertain to various domains, and are accompanied by aliases and descriptions that greatly aid annotators. Since many references in Wikidata can be automatically verified through API calls, as showcased by Amaral et al.~\cite{amaral2021assessing}, WTR is built focusing on those that can not. Furthermore, to prevent biases towards frequently used web domains, such as Wikipedia and The Peerage, WTR is built to represent a variety of web domains with equal amounts of samples from each.

\subsubsection{Selecting References}

WTR is constructed from the Wikidata dumps from March 2022. First, all references are extracted. Those associated to at least one triple and which lead to a web page by either an external URL (through the \texttt{reference URL (P854)} property), or by an external identifier property that has formattable URLs, e.g. \texttt{VIAF ID (P214)}, are kept. These two types of references constitute $91.52\%$ of all references. The remaining portion consists of references to Wikidata items, inferred from other specific Wikidata claims, imported from Wikipedia (but no page specified), and those without any provenance property. These types of references are avoided due to potential issues with circular or vague provenance. Close to 20M unique references are extracted.

Next, each extracted reference has their initial web URLs defined. For references with direct URLs (\texttt{reference URL (P854)}), such URLs are used. For references with external ID properties, the property's \texttt{formatter URL (P1630)} is combined with the linked external ID to establish the URL. For instance, a reference might use the \texttt{IMDb ID (P345)} property, which has the formatter URL \url{``https://wikidata-externalid-url.toolforge.org/?p=345\&url\_prefix=https://www\\.imdb.com/\&id=\$1''}. By replacing the `$\$1$' with the ID linked by the property, one establishes the IMDB URL represented by the reference.

The extracted set of references and their respective initial URLs is then filtered. References that are inadequate to the scenario in which ProVe will be used are removed. These consist of three groups:

\begin{enumerate}
    \item References with URLs to domains that have APIs or content available as structured data (e.g. JSON or XML), as these can be automatically checked through APIs, e.g. PubMed, VIAF, UniProt, etc.
    \item References with URLs linking to files such as CSV, ZIP, PDF, etc., as parsing these file formats is outside of ProVe's scope.
    \item References with URLs to pages that have very little to no information in textual format, such as images, document scans, slides, and those consisting only of infoboxes, e.g. \url{nga.gov}, \url{expasy.org}, Wikimedia commons, etc.
\end{enumerate}

As shown by Amaral et al.~\cite{amaral2021assessing}, this first group is very substantial, with an estimated over $70\%$ of English references being automatically verifiable through API calls. Additionally, references with URLs to websites not in English (according according to FastText's language identification models~\cite{joulin2016fasttext,joulin2016bag}), posing security risks, or unavailable (e.g. 404 and 502 HTML response codes) are removed. 

Close to 7M references are left after these removals, wherein English Wikipedia alone represents over $40\%$ of URLs. To avoid biasing evaluation towards populous web domains, a stratified sampling is carried using web domains as strata. The $30$ most populous web domains are defined as $30$ separate groups, with two additional groups formed from remaining web domains: \textit{RARE}, for web domains that appear only once, and \textit{OTHER}, for all others. 

\subsubsection{Pairing References with Triples}

Given the total number of references contemplated (7M), a sample size of $385$ represents a $95\%$ confidence interval and a $5\%$ margin of error. An equal amount of references from each of the $32$ groups is sampled, totalling over $400$ references. Samples for a group are retrieved one by one through the following method. 
A reference is randomly sampled without replacement and checked as to whether it is associated to at least one triple fitting to be used for evaluation; if so, it is kept. A triple fitting for evaluation is defined as one which carries non-ontological and reliable meaning, and which can be expressed concisely through natural language. They are identified by the following criteria:

\begin{itemize}
    \item Is not deprecated by Wikidata;
    \item Has an object that is not the ``novalue'' or ``somevalue'' special nodes;
    \item Has an object that is not of an unverbalisable type, i.e. URLs, globe coordinates, external IDs, and images.
    \item Has a predicate that is not ontological, e.g. \texttt{instance of (P31)}, \texttt{merged into (P7888)}, \texttt{entry in abbreviations table (P8703)}, etc.
\end{itemize}

These steps produce a stratified representative sample of references, including their unique identifiers, resolved URLs, web domains, and HTTP response attributes. Finally, for each sampled reference, a triple-reference pair is formed by extracting from Wikidata a random triple associated to the reference and fitting for evaluation, alongside its unique identifiers, object data types, main labels, aliases, and descriptions. This construction process creates WTR, ensuring it is composed of triples carrying non-ontological meaning and verifiable through their associated references, thus useful for evaluating ProVe. It also ensures meaning and context understanding is evaluated, rather than merely string matching, e.g. in the case of URLs, globe coordinates, and IDs. As for image data, tackling such multimodal scenarios is outside ProVe's scope. 

\subsection{Dataset Annotation}
\label{sec:data_ann}

%Elena: Add task screenshots in the appendix. Add sub-sections for tasks, evaluation, recruiment etc. Mention ethics approval.

% Move screenshots to appendix
%Move final example of dataset format to the appendix
%Add subsections for task, evaluation, recruitment, etc (look into datasheets for crowdsourcing)
%Make sure to quote ethics approval

As described in Section~\ref{sec:method} (see Figure~\ref{fig:example}), ProVe tackles its AFC task as a sequence of text extraction, ranking, classification, and aggregation subtasks. Given a triple-reference pair, ProVe extracts text passages from the reference, ranks them according to relevance to the triple, and individually classifies them according to their stance towards the triple. Then, triple-reference pairs are classified according to the overall stance of the reference towards the triple.

To allow for a fine-grained evaluation of ProVe's subtasks, WTR receives three sets of annotations: (1) on the stance of individual pieces of evidence towards the triple, (2) on the collective stance of all evidence, and (3) on the overall stance of the entire reference. The two first sets of annotations are deemed evidence-level annotations, while the last is reference-level. Crowdsourcing is used to collect evidence-level annotations, due to the large number of annotations needed (six per triple-reference pair) in combination with the simplicity of the task, which requires workers only to read short passages of text. Reference-level annotations are less numerous (one per triple-reference pair), much more complex, and hence manually annotated by the authors.

%To collect the necessary annotations, we rely on both crowdsourcing and manual annotations by the authors. Firstly, crowdsourcing is used to collect annotations at the sentence levels (first two levels), due to the large number of annotations needed in combination with the simplicity of the task, which requires workers only to read short passages of text. As for annotations on the overall stances of referenced web sites, i.e. at reference level, those are less numerous, much more complex, and hence manually annotated by the authors.

\subsubsection{Crowdsourcing Evidence-level Annotations}
Collecting evidence-level annotations for all retrievable sentences of each triple-reference pair in WTR, in order to account for different rankings that can be outputted by ProVe, would be prohibitively expensive and inefficient. Thus, evidence-level annotations are only provided for the five most-relevant passages in each reference, i.e. the collected evidence. First, ProVe's text retrieval (Section~\ref{sec:method_text}) and sentence selection (Section~\ref{sec:method_sensec}) modules are applied to each reference and the five pieces of evidence for each collected. This does not severely bias the annotation towards highly-relevant text passages, actually allowing for a more even collection of both relevant and irrelevant text passages, as often only a couple of passages amidst the five tend to be relevant. Then, evidence-level annotations for each individual piece of evidence, as well as for the whole evidence set, are collected through crowdsourcing, totalling $6$ annotation tasks per triple-reference pair.

\paragraph{Task Design}
Two crowdsourcing task designs have been created to carry out this evidence-level annotation process. Task design 1 (T1) asks workers to assess the stance of an individual evidence towards the triple, which can also be used as an indication of relevance. Each task in T1 is a bundle of $6$ subtasks, each providing the worker with a Wikidata triple, a piece of evidence extracted from its paired reference, the reference's URL, and asking the worker the stance of that evidence: either it \textbf{supports} the claim, \textbf{refutes} it, has \textbf{not enough information} for either, or the worker is \textbf{not sure}. This bundling is done in order to get more annotations out of a single crowdsourcing task assignment. Task design 2 (T2) asks workers to assess the collective stance of the evidence set (all five individual pieces of evidence) towards the triple. Similarly to tasks in T1, tasks in T2 are made from $6$ subtasks. Each T2 task providing the worker with a triple, five text passages extracted from the reference (the evidence set), shown in a random order, and the reference's URL. It then asks the \textit{collective} stance of the five passages, with similar response options to T1. The designs for both T1 and T2 tasks can be seen in Appendix~\ref{appendix:1}.

\paragraph{Recruitment and Quality Assurance}
The crowdsourcing campaign received ethical approval by King's College London on 7th of April, 2022, with registration number MRA-21/22-29584. All tasks were carried through Amazon Mechanical Turk (AMT). A pilot was run to collect feedback on instructions, design, and compensation. After proper adjustments, the main data annotation tasks were carried out. Several quality control techniques~\cite{daniel2018quality} were applied, following similar tasks by Amaral et al.~\cite{amaral2022wdv,amaral2021assessing}. A number of subtasks in both T1 and T2 were manually created and annotated by the authors and used as golden-standard subtasks to reject low-quality workers. A randomised attention test was put at the start of each task to discourage spammers. Finally, detailed instructions and examples were available at all times to workers.

Execution times for each task in the pilot were measured and used to define payment in USD proportionally to double the US minimum hourly wage ($7.25$): USD $0.50$ for tasks in T1 and USD $1.00$ for tasks in T2, calculated based on the highest between mean and median execution time. $500$ tasks were generated for T1 and $91$ tasks for T2, assigned to about $200$ and $140$ unique workers, respectively. Workers needed to have finished at least $1000$ tasks in AMT with at least $80\%$ approval. Each task was resolved $5$ times and annotation aggregation was done via majority voting to reduce worker bias. In the event of a tie ($4\%$ of cases for T1 and $13.4\%$ for T2), authors served as tie-breakers.

To assure annotation aggregation was trustworthy, inter-annotator agreement was measured through kappa values, achieving $0.56$ for tasks in T1 and $0.33$ for tasks in T2. According to Landis and Koch~\cite{landis1977measurement}, these results show fair and moderate agreement, respectively. Several factors that contribute to lower inter-annotator agreement~\cite{bayerl2011determines} are present in this crowdsourcing setting: subjectivity inherent to natural language interpretation, a high number of annotators which also lack domain expertise, and class imbalance. On individual annotations for T1, the majority of passages ($68.7\%$) was deemed as neither supporting nor refuting, followed by passages annotated as supporting ($27.5\%$), and a small portion refuting ($3.3\%$). Only $0.4\%$ of annotations were `not sure'. Aggregated by majority-voting, these values are, respectively, $70.5\%$, $28.5\%$, $1.0\%$, and $0.0\%$. For T2, the proportions of individual (and aggregated) annotations were $65.5\%$ ($73.6\%$) for supporting sets of evidence, $9.3\%$ ($5.9\%$) for refuting, $24.6\%$ ($20.5\%$) for neither, and $0.6\%$ ($0.0\%$) for `note sure'.

\subsubsection{Gathering Reference-level Annotations}

WTR has reference-level annotations for each triple-reference pair. They define a reference's overall stance towards its associated triple, and are manually provided by the authors. These annotations are crucial in order to provide a ground truth for an evaluation of the entire pipeline's performance when taking the whole web page into consideration. They consider a reference's full meaning and context, and not only what was captured and processed by the modules as evidence. Differently from the other annotation level, covered by crowdsourcing tasks that could be simplified to directly comparing extracted text passages, the mental load and task complexity of interacting with the page to inspect all information (e.g. in text, infoboxes, images, charts), on top of cases where the information is nonexistent, is too high for cost-effective crowdsourcing. Thankfully, with one annotation per triple-reference pair, it is feasible for manual annotations to be created by the authors. The authors have thus annotated the over $400$ references into different categories and sub-categories, which are a more detailed version of the three stance classes used at evidence-level annotations (and by FEVER):

\begin{enumerate}
    \item Supporting References (directly maps to the `\textbf{supports}' class):
    \begin{enumerate}
        \item Support explicitly stated as text as natural language sentences
        \item Support explicitly stated as text, but not as natural language sentences
        \item Support explicitly stated, but not as text
        \item Support implicitly stated
    \end{enumerate}
    \item Non-supporting References
    \begin{enumerate}
        \item Reference refutes claim (directly maps to the `\textbf{refutes}' class)
        \item Reference neither supports nor refutes the claim (directly maps to the `\textbf{not enough information}' class)
    \end{enumerate}
\end{enumerate}

These six subclasses allow WTR to aid in evaluating the overall performance of ProVe in both ternary (the three sentence-level stances) and binary (supporting vs. not supporting) classifications, as well as to investigate which presentations of supporting information are better captured by the pipeline.

WTR contains $416$ Wikidata triple-reference pairs, representing $32$ groups of text-rich web domains commonly used as sources, as well as $76$ distinct Wikidata properties. $43\%$ of references were obtained through external IDs and $57\%$ through direct URLs. Its structured is shown in Appendix~\ref{appendix:2}.
\section{Evaluation}
\label{sec:eval}

%Elena: Again, you need sub-sections in 5.1, 5.2, 5.3, and 5.4 because you're essentially evaluating four different tasks and it is a lot to take in. For each task have a secion on what you evaluate, metrics, summary of results. Add formulas for all metrics used.

%Subsections:
%- Validation on dev/test data
%- What is evaluated on WTR
%- Metrics
%- Summary of results

This section covers the evaluation of ProVe's performance by applying it to the evaluation dataset WTR, described in Section~\ref{sec:data}, and comparing ProVe's outputs with WTR's annotations. These inspections and comparisons provide insights into the pipelines' execution and results at its different stages and modules. Each module in ProVe is covered in a following subsection. The overall classification performance of ProVe is indicated by the outputs of the claim verification module's aggregation step and is covered at the end of the section.

\subsection{Claim Verbalisation}
%verbalisation
\label{sec:eval_claimverb}

Given that ProVe's verbalisation module is the exact same model used to create the Wikidata triple verbalisations found in the WDV dataset~\cite{amaral2022wdv}, this section first reports the relevant evaluation results obtained by WDV's authors. It then analyses the claim verbalisation module's execution on the WTR evaluation dataset, looking at the quality of its outputs.

\subsubsection{Model Validation}
ProVe's verbalisation module consists of a pre-trained T5 model~\cite{raffel2020exploring} fine-tuned on the WebNLG dataset~\cite{gardent2017webnlg}. To confirm that fine-tuning was properly carried, the authors measure the BLEU~\cite{papineni2002bleu} scores of its verbalisations on WebNLG data. They measure $65.51$, $51.71$, and $59.41$ on the testing portion of the \textit{SEEN} partition, on the \textit{UNSEEN} partition, and on their combination, respectively. These are all within a percentage point from current state-of-the-art~\cite{ribeiro2020investigating}.

Amaral et al.~\cite{amaral2022wdv} use this exact same fine-tuned model to create multiple Wikidata triple verbalisations, which compose the WDV dataset, and evaluate them with human annotators. WDV consists of a large set of Wikidata triples, alongside their verbalisations, whose subject entities come from three distinct groups (partitions) of Wikidata entity classes. The first partition consists of 10 classes that thematically map to the 10 categories in WebNLG's \textit{SEEN} partition. The second, of 5 classes that map to the 5 categories in WebNLG's \textit{UNSEEN} partition. The third consists of 5 new classes not covered in WebNLG but populous in Wikidata.

WDV's verbalisations were evaluated by Amaral et al. through aggregated crowdsourced human annotations of fluency and adequacy dimensions, as defined in Section~\ref{sec:method_verb}. Fluency scores range from $0$, i.e. very poor grammar and unintelligible sentences, to $5$, i.e. perfect fluency and natural text. Adequacy scores consisted of \textit{0/No} for inadequate verbalisations and \textit{1/Yes} for adequate ones. WDV's authors observed $96\%$ of annotated verbalisations having a median fluency score of 3 or higher, where 3 denotes ``Comprehensible text with minor grammatical errors'', and around $93\%$ being voted by the majority of annotators as adequate. These results did not vary considerably between WDV partitions, indicating model stability regardless if classes are seen in training or have mappings to WebNLG (DBpedia) classes. The WDV paper~\cite{amaral2022wdv} contains a more detailed evaluation.

\subsubsection{Execution on WTR}
The verbalisation module was applied to all $416$ triple-reference pairs in WTR. For each triple $t$, its three components were retrieved from Wikidata: subject ($s$), predicate ($p$), and object ($o$). Subjects and predicates are all Wikidata entities and, thus, have associated main labels and aliases. Objects have multiple possible data types, including Wikidata entity, from which labels were retrieved like with subjects. Strings and quantities without units are used as-is as single labels, otherwise the unit's main label and aliases are added. Date-time values are formatted into multiple string patterns based on their granularity. 

The process of defining preferred labels for verbalisation, represented in Section~\ref{sec:method_verb} through function $l(\cdot)$, consisted of using the main labels of all three components for verbalisation, and changing the predicate's label for an alias in case the verbalisation ($v$) was not fluent or adequate. First, for a triple $t$, a concatenation of its components' main labels is fed to the verbalisation model to generate a verbalisation $v$. Then, these verbalisations were manually inspected by the authors in order to assess fluency and adequacy as previously defined. In case a verbalisation $v$ scored lower than a $4$ on fluency or was inadequate, it was replaced by an alternate verbalisation $v'$ generated by using a predicate alias rather than the predicate's main label. As mentioned in Section~\ref{sec:method_claimver}, contextually-dependant and vague predicate main labels hinder verbalisations and choosing proper aliases for it can be quickly carried out by KG editors and curators. One example is the predicate \texttt{child (P40)}, whose alias `has child' is used to remedy the main label's lack of explicit direction.

Out of the $416$ verbalisations generated by ProVe through main labels, $62.6\%$ were adequate and had good to excellent fluency. Another $29.8\%$ followed suit after predicate alias replacement, totalling $92.4\%$. The remaining $7.6\%$ either had no aliases or could not be improved by them and had to be manually corrected before being passed down the pipeline. While manual corrections are more demanding than alias selection, those were not frequent and mainly affected specific properties such as \texttt{isomeric SMILES (P2017)} and \texttt{designation (P1435)}. Corrections were also not extensive; the normalised Levenshtein distance before and after corrections was under $0.25$ (a quarter total length) for $80\%$ of them.

Finally, $7$ of the claims verbalised ended up having both identical URLs and verbalisations, and were thus dropped from the evaluation downstream as they would have the exact same results. This results from ProVe not taking claim qualifiers into consideration, which is further discussed in Section~\ref{sec:disc_quali}.

\subsection{Text Extraction}
\label{sec:eval_text}

Due to the complexity of defining metrics that measure success in text extraction, this section instead first defines metrics that can be used for an indirect evaluation of the text extraction module. It then explores insightful descriptive metrics obtained from executing the module on WTR.

\subsubsection{Indirect Evaluation}

ProVe's text extraction module essentially performs a full segmentation of the referenced web-page's textual content without excluding any text. The model can not be directly evaluated through annotations due to the sheer quantity of ways in which one can segment all references contained in the evaluation dataset. Annotating such text extractions would require manually analysing entire web pages to find all textual content relevant to the claim, inspecting all the text extractable from such page, and segmenting it such that boundaries are properly placed in terms of syntax and that relevant passages are kept unbroken. One would then need to compare one or multiple ideal extractions to the extraction performed by ProVe. It is neither trivial to simplify this process for crowdworkers, nor efficient for the authors to carry it out by hand. It is also not trivial to define what constitute `well-placed' sentence boundaries nor how and if one can break relevant passages of text. 

Thus, instead of a direct evaluation, the performances of the subsequent sentence selection and final aggregation steps are used as indirect indicators of ProVe's text extraction. A correlation between ProVe's relevance scores ($\rho$) and evidence-level crowd annotations, as defined in Section~\ref{sec:eval_sentselect_correlation}, can be used to measure how much ProVe captures human-perceived relevance. A high value of such metric indicates ProVe extracts text passages such that relevant and irrelevant text segments are well divided. Otherwise, there would be a dissonance between humans and models in rating relevance, as while humans are good in judging badly divided passages, models would struggle significantly.

%Mention high classification performance here as a metric and define it (reference to sentence selection section, also to equation X)
Likewise, a good final classification performance, measured against WTR's reference-level annotations, indicates ProVe's capacity of extracting useful sentences. Classification metrics for ternary and binary classification tasks, such as accuracy and f-scores, are shown in Section~\ref{sec:eval_claimverification}. Still, the direct evaluation of ProVe's text extraction module, encompassing sentence segmentation and meaning extraction from unstructured and semi-structured textual content, is intended as future work. 

\subsubsection{Execution on WTR}

The text extraction module was applied to each of the $416$ triple-reference pairs $(t,r)$ in WTR, each yielding a set of passages $P$, as defined in Section~\ref{sec:method_text}. The total number of extracted passages was of nearly $64$K, an average of $154$ passages per reference. This average number of extracted passages varied heavily according to web domain, ranging from as low as $1$ to as high as $804$, with relatively low inter-domain standard deviation ($29.51$ median standard deviation). The average size, in characters, of individual extracted passages behaved similarly, ranging from $24.20$ to $4059.23$ depending on web domain. Finally, the number of passages $|P|$ extracted from a reference $r$ and their average size $(\sum_{i=0}^{|P|-1} |\mathtt{p_{i}}|) / |P|$ have a slightly moderate negative correlation ($-0.2513$ Pearson's r), with a few domains, such as \url{bioguides.congress.gov} returning very few but very large passages.

These metrics confirm that extracted textual content mainly vary based on web domain. It indicates ProVe's extraction depends heavily on particular web layouts, e.g. having difficulty segmenting the contents of specific domains like \url{bioguides.congress.gov}, due to their textual content being contained in a single paragraph ($<$p$>$ HTML tag) without periods or any other sentence breaks.

\subsection{Sentence Selection}
\label{sec:eval_sentsec}

ProVe's sentence selection module contains a BERT model fine-tuned on FEVER's training partition, as described in Section~\ref{sec:method_sensec}. This section first performs a sanity check of by evaluating the model on FEVER's validation and testing partitions, measuring standard classification metrics to ensure the model has properly fine-tuned to FEVER. Afterwards, the entire module is applied to WTR and its performance measured by relying on the crowdsourced annotations.

\subsubsection{Model Validation}

For each claim-sentence pair in FEVER's validation and testing partitions, the sentence selection module's BERT model outputs a relevance score between $-1$ and $1$. A sentence is classified as relevant to a claim if it figures within the top 5 highest scoring sentences for that claim, and as irrelevant otherwise. According to FEVER Scorer~\footnote{\url{https://github.com/sheffieldnlp/fever-scorer}}, the model scores $0.945$ and $0.87$ recall on validation and testing, respectively. Similarly, F1-scores were $0.421$ and $0.386$, which are less than a percentage point away from KGAT~\cite{liu2019kgat} and DREAM~\cite{zhong2019dream}, current state-of-the-art, on the test partition.\footnote{\url{https://competitions.codalab.org/competitions/18814\#results}} These indicate the model has properly fine-tuned to FEVER.

\subsubsection{Execution on WTR}

Inputs to the sentence selection module consist of the verbalisations $v$ outputted by the claim verbalisation module, as described in Section~\ref{sec:eval_claimverb}, and the extracted passages $P$ taken from the reference, as described in Section~\ref{sec:eval_text}. For each passage $\mathtt{p} \in P$, a relevance score $\rho \in [-1,1]$ is calculated.

\paragraph{Distribution of Relevance Score}

\begin{figure}[h]
  \centering
  %\vspace*{-15pt}
  \includegraphics[width=0.9\linewidth]{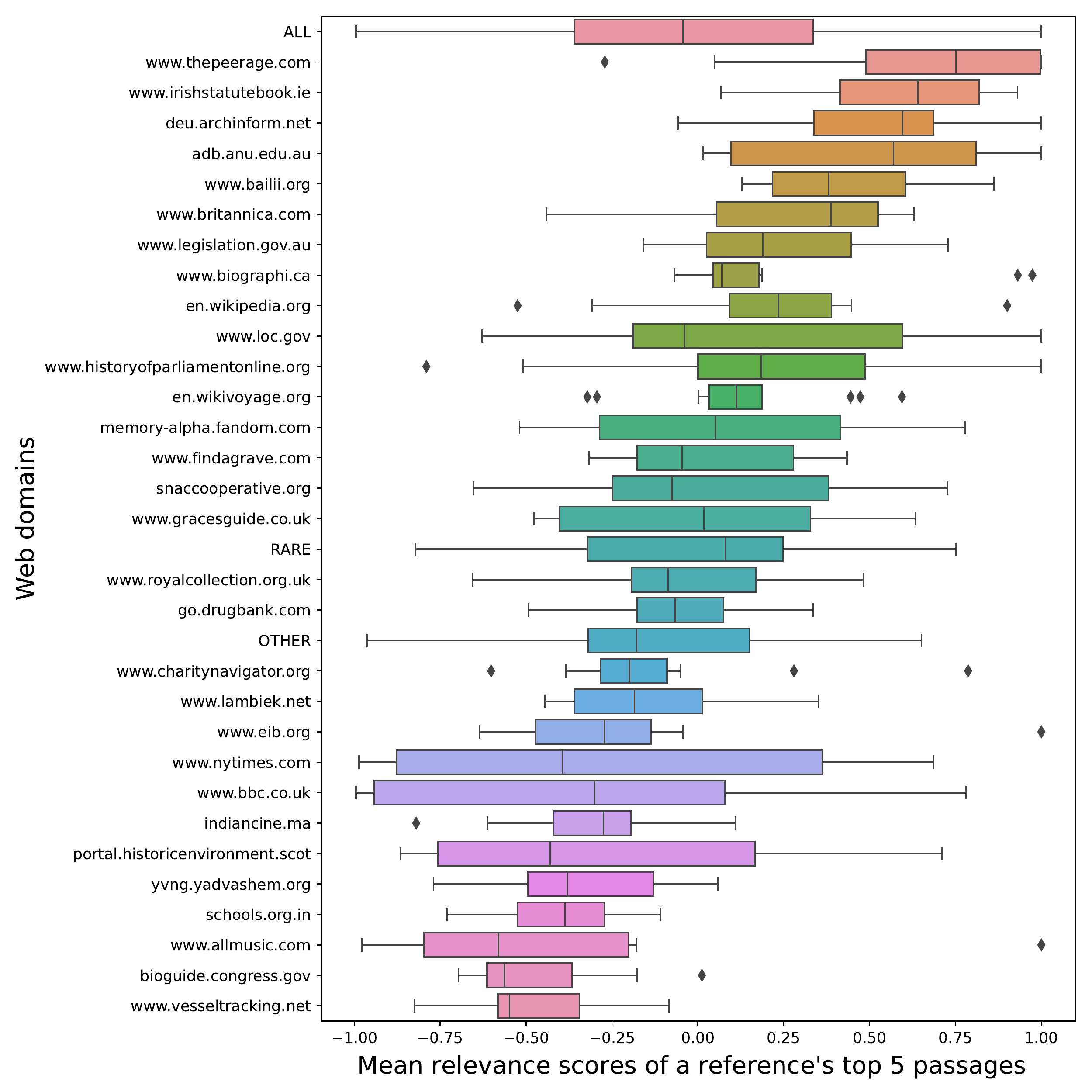}
  %\vspace*{-15pt}
  \caption{Relevance scores distributions across and within different web domains. `ALL' stands for the combined distribution of the subsequent 32 groups. Data values here are the averages of the top $5$ passages' relevance scores for each reference.}
  \label{fig:relevance_per_domain}
%\vspace*{-50pt}
\end{figure}

Relevance scores ($\rho$) varied heavily across web domains. For each triple-reference pair $(t,r)$ in WRT, its passages $P$ were scored by the sentence selector module, and the top $5$ highest scores (which correspond to the evidence set $E$) were averaged, with Figure~\ref{fig:relevance_per_domain} showing the distribution of these averages across and within domains. Overall, relevance scores spanned the whole range of values (-1 to 1), with a median close to zero. Variation within web domains was not wide, denoted by a prevalence of small to medium interquartile ranges. Web domains with large variations were those that cover a large range of information domains and page layouts, such as \url{bbc.co.uk}. In contrast, there is extensive variation across web domains. This is due, firstly, to how each domain conveys its information, where long and continuous textual contents greatly favour reliable scores, as opposed to spread-out information which pushes scores down. Secondly, due to the amount of content actually related to the triple. A website like \url{thepeerage.com} mentions the triples's subject many times, leading to many positively-scored sentences, as opposed to \url{vesseltracking.net}, which provides fewer, shorter, and more direct information.

\paragraph{Proportion of Irrelevant Passages}

Given the distributions seen in Figure~\ref{fig:relevance_per_domain}, one can define the value zero as the threshold between likely relevant and likely irrelevant passages. By using only the passages extracted by the $n=1$ sliding window ($P_1$), this threshold leaves $24.7\%$ of triple-reference pairs as containing only passages that are likely irrelevant. This decreases to $17.6\%$ when combining the $n=1$ and $n=2$ sliding windows, clearly indicating the method described in Section~\ref{sec:method_text} achieves its desired objective of generating more relevant passages by combining multiple text segments. This percentage of triple-reference pairs without likely relevant passages also varies heavily across web domains, and is a problem mostly affecting those same domains at the lower end of the distribution seen in Figure~\ref{fig:relevance_per_domain}.

\paragraph{Correlation Between Modeled and Crowdsourced Relevance}
\label{sec:eval_sentselect_correlation}

For each triple-reference pair $(t,r)$ in WTR, crowdworkers annotated each of its $5$ most relevant passages (its Evidence set $E$) as ranked by the sentence selection module, described in Section~\ref{sec:data_ann} as the evidence-level annotations obtained through task design T1. Pieces of evidence were annotated according to their individual stance towards the reference's associated triple, using one of four choices: `supports', `refutes', `neither' (not enough information), and `not sure'. In order to reduce crowd bias, multiple individual annotations were collected per evidence and aggregated through majority voting. Authors served as tie-breakers.

\begin{figure}[]
  \centering
  \includegraphics[width=1\linewidth]{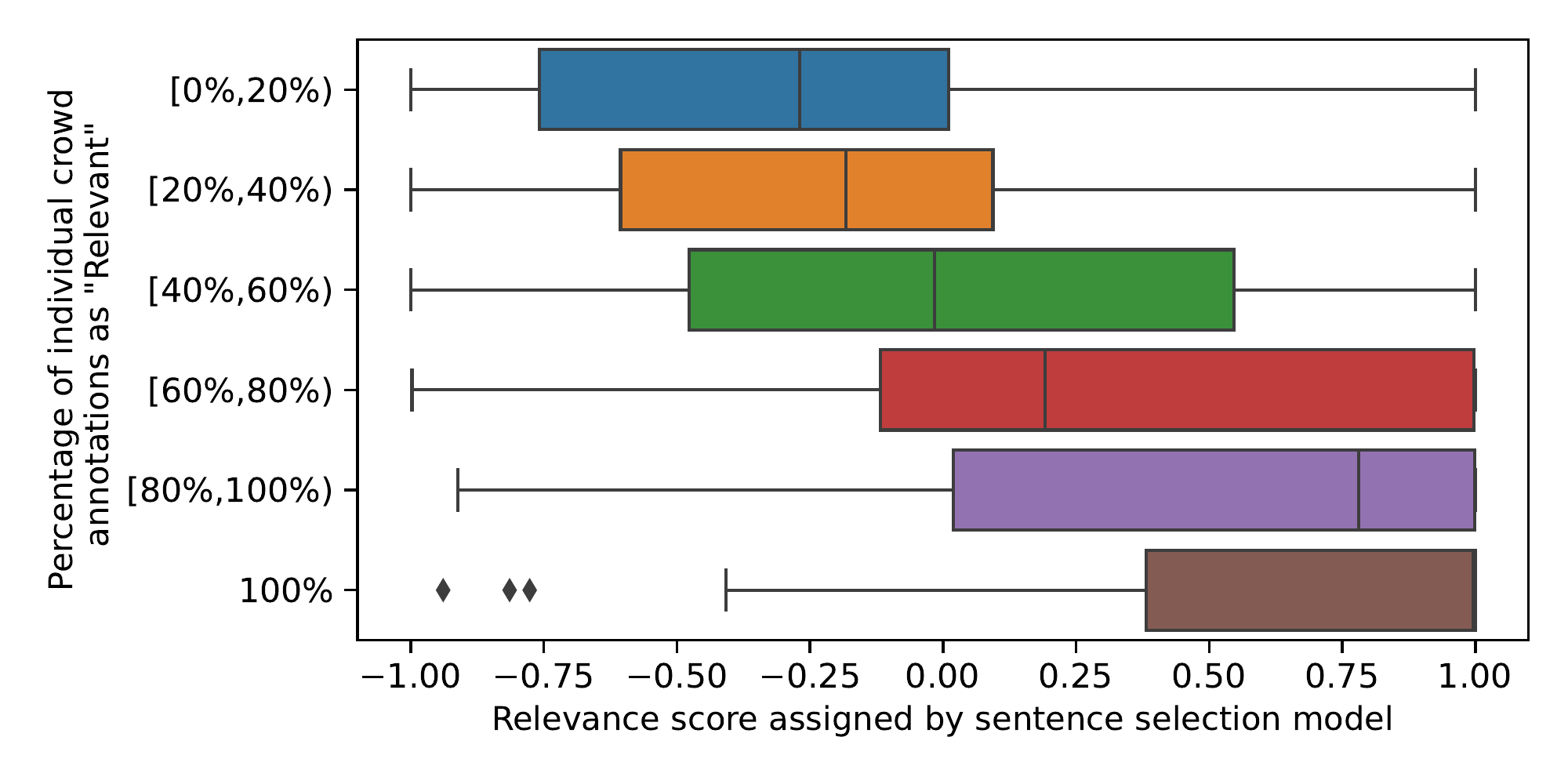}
  \caption{Distributions of relevance scores given by the module divided by the percentage of crowd annotations deeming that passage as `relevant' (either `supports' or `refutes'.}
  \label{fig:relevance_per_annotation_per}
\end{figure}

In order to compare the model's relevance scores ($\rho$) with human-perceived values of relevance (T1 crowd annotations), both `supports' and `refutes' annotations are grouped as `relevant', with `neither' relabelled as `irrelevant'. Relevance score distributions are then analysed based on this binary class annotation. Figure~\ref{fig:relevance_per_annotation_per}, shows annotated passages divided into groups based on the percentage of `relevant' individual annotations (or votes) each has received, as well as each group's relevance score distribution. There is a clear pattern of association between the model's relevance scores and real humans' opinions of relevance. This conclusion is reinforced by the strong correlation between relevance scores and percentages of individual annotations as `relevant' ($0.5058$ Pearson's r).

Such strong correlation can also be seen with aggregated annotations, as shown in Figure~\ref{fig:relevancy_score_distributions_kde}, where the relevance score distributions for the group of passages majority-voted as `relevant' vs. those voted as `irrelevant'. These metrics and distributions indicate that ProVe's sentence selection module produces scores that are well-related to human judgements of relevance.

\subsection{Claim Verification}
\label{sec:eval_claimverification}
%ter/claim verification

The first step of ProVe's claim verification module consists of a TER classification task, resolved by a fine-tuned BERT model, as described in Section~\ref{sec:method_claimver}. Such TER model is used to classify the stances of individual pieces of evidence ($e \in E$) towards a KG triple ($t$) by calculating three class probabilities ($\sigma$) corresponding to the three TER classes found in FEVER: `supports', `refutes', and `not enough information'. The module's second step is an aggregation which uses these classification probabilities to calculate a final verdict for the whole reference ($r$). This section first describes a sanity-check of both steps by evaluating them in conjunction on FEVER's validation and testing partitions. It then assesses their performance on WTR.

\subsubsection{Model Validation}

ProVe's claim verbalisation module has been applied to FEVER's validation and testing partitions. For the validation partition, the sentences pre-annotated as relevant to the claim being judged were used as evidence. The first TER step is performed to calculate the individual stance probabilities of each piece of evidence. The second aggregation step is then carried out to define the claim's final verdict. The same process is carried for the testing partition, however, since its sentences do not come pre-annotated, ProVe's sentence selection module was used instead.

For the final aggregation step, only methods $\#1$ (a weighted sum) and $\#2$ (Malon's strategy) were used, as neither require training a new classifier and thus have similar complexity to other approaches used to tackle FEVER~\cite{malon2019team,soleimani2020bert}, yielding a more direct comparison. Label accuracy and FEVER score were calculated as evaluation metrics. Label accuracy is a normal classification accuracy calculated over the three TER classes. FEVER score is an accuracy calculated by also taking collected evidence into consideration: a prediction is correct if the label is correct \textbf{and} the predicted evidence set contains all correct evidence.

Aggregation method $\#1$, the weighted sum, scores $0.6964$ label accuracy and $0.6952$ FEVER score on the validation set, and $0.6508$ and $0.617$ respectively on the testing set. Method $\#2$, the rule-based aggregation, scores $0.7624$ label accuracy and $0.76110$ FEVER score on the validation set, and $0.7037$ and $0.6739$ respectively on the testing set. Method $\#2$ puts ProVe $3.2$ percentage points below state-of-the-art (DREAM~\cite{zhong2019dream}) on FEVER score, which the authors consider sufficient to validate the module's fine-tuning. 

\subsubsection{Execution on WTR}

ProVe's claim verification module is tested on WTR by comparing its outputs to WTR's annotations, both at evidence level and reference level. Such annotations, as described in Section~\ref{sec:data_ann}, consist of: crowd annotations denoting the stances of individual pieces of evidence towards KG triples (from crowdsourcing tasks T1), crowd annotations denoting the collective stances of sets of evidence towards KG triples (from crowdsourcing tasks T2), and author annotations denoting the stance of the entire reference towards its associated KG triple. Crowdsourced annotations are collected multiple times and aggregated through majority voting, with authors serving as tie-breakers.

\paragraph{Distributions of TER Class Probabilities}

Similarly to the relevance scores outputted by the sentence selection module, the TER class probabilities outputted by the TER model for individual pieces of evidence varied greatly across web domains. Consider a triple-reference pair $(t,r)$ and its $i$-th piece of evidence $e_i$, whose stance probabilities assigned by the model are $(\sigma^{SUPP}_{i}, \sigma^{REF}_{i}, \sigma^{NEI}_{i})$. Consider also the highest of the three probabilities to denote the predicted TER class for that individual evidence $z_i$, where $z_i = \argmax_{k \in \mathcal{K}} \sigma^k_i$. Web domains such as \url{indiancine.ma} have close to $95\%$ of its evidence classified as `not enough information', while domain such as \url{deu.archinform.net} classify close to $98\%$ as `supports'. Likewise to sentence selection, such variations can be attributed to web layouts. For instance, infoboxes, implicit subjects, and large amounts of boilerplate clearly inflate the number of unrelated sentences that might get highly ranked for relevance, but are neither supportive nor refutative.

\paragraph{Individual TER Classification Metrics}

\begin{figure}[]
\begin{minipage}{0.49\linewidth}
\includegraphics[width=1\linewidth]{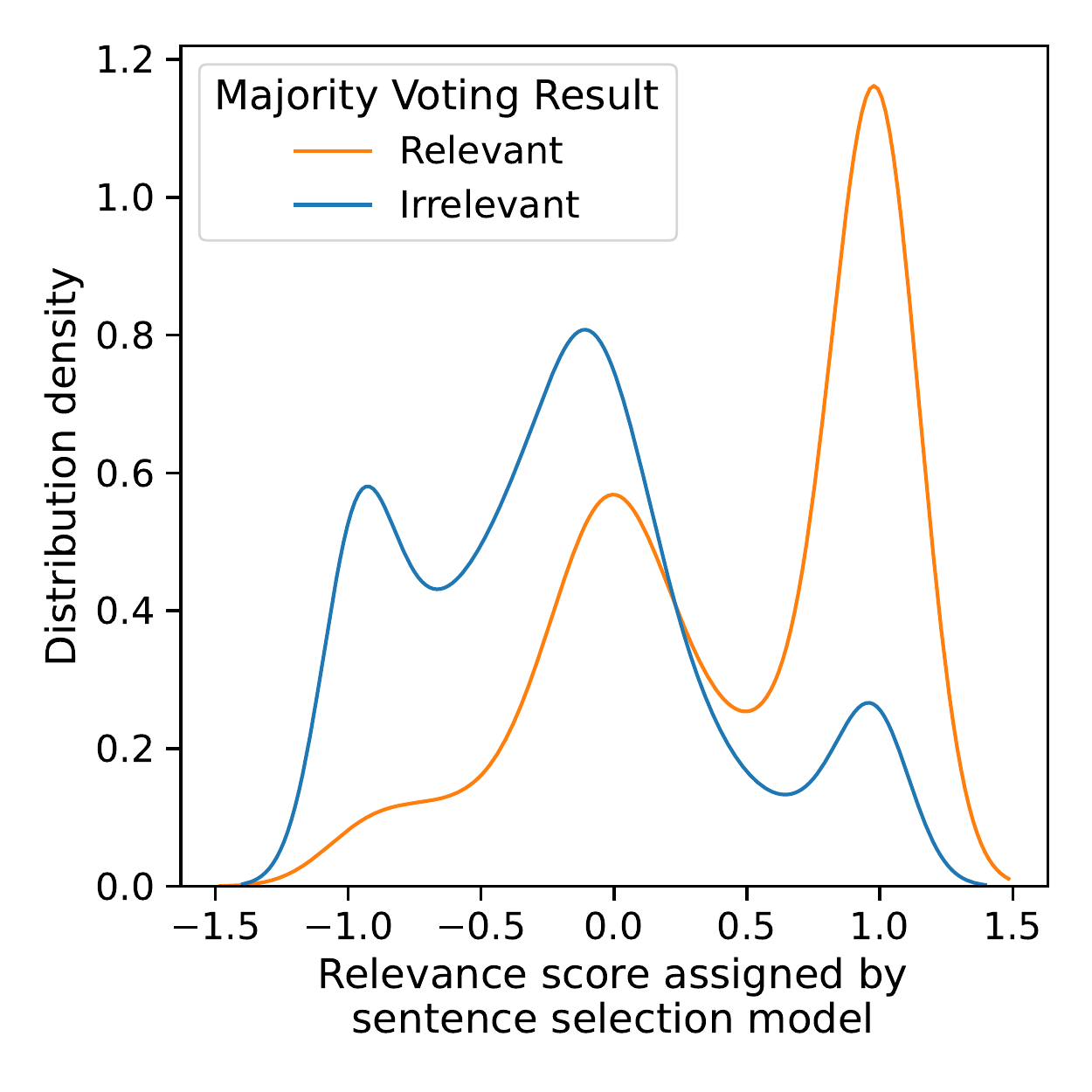}
\caption{Relevance score distributions of passages majority-voted as `relevant' and of those voted `irrelevant'.}
\label{fig:relevancy_score_distributions_kde}
\end{minipage}
\hfill
\begin{minipage}{0.49\linewidth}
\includegraphics[width=1\linewidth]{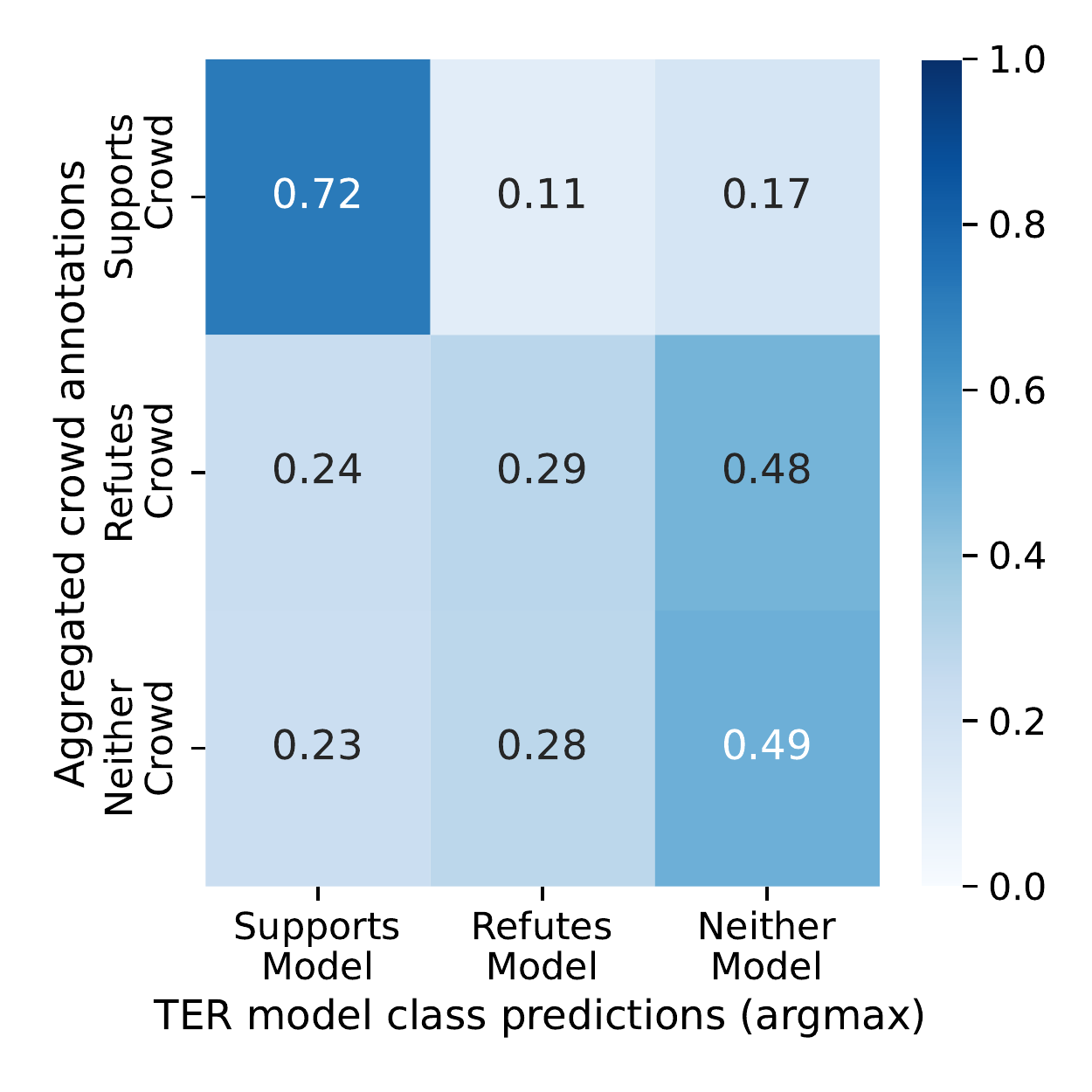}
\caption{Stance classes predicted for single claim-reference pairs (obtained through argmax) vs. the crowd's aggregated annotations.}
\label{fig:single_evidence_agg_CM}
\end{minipage}
\end{figure}

Using this argmax approach to define the predicted stance $z_i$ of an individual piece of evidence $e_i$, and using the aggregated annotations provided by the crowd through T1 tasks as ground truth, once can measure classification metrics for ProVe's individual TER classification. Figure~\ref{fig:single_evidence_agg_CM} shows the resulting confusion matrix. Accuracy was $0.56$ and Macro F1-score was $0.43$. While predicting supporting sentences with moderate precision, ProVe's TER model has difficulty in disentangling refuting sentences from those that neither support nor refute. This is believed to be due to the differences between refuting evidence data found in FEVER and refuting references that occur naturally in Wikidata's sources, discussed in more detail in Section~\ref{sec:disc_ref}.

\begin{figure}[]
\begin{minipage}{0.49\linewidth}
\includegraphics[width=1\linewidth]{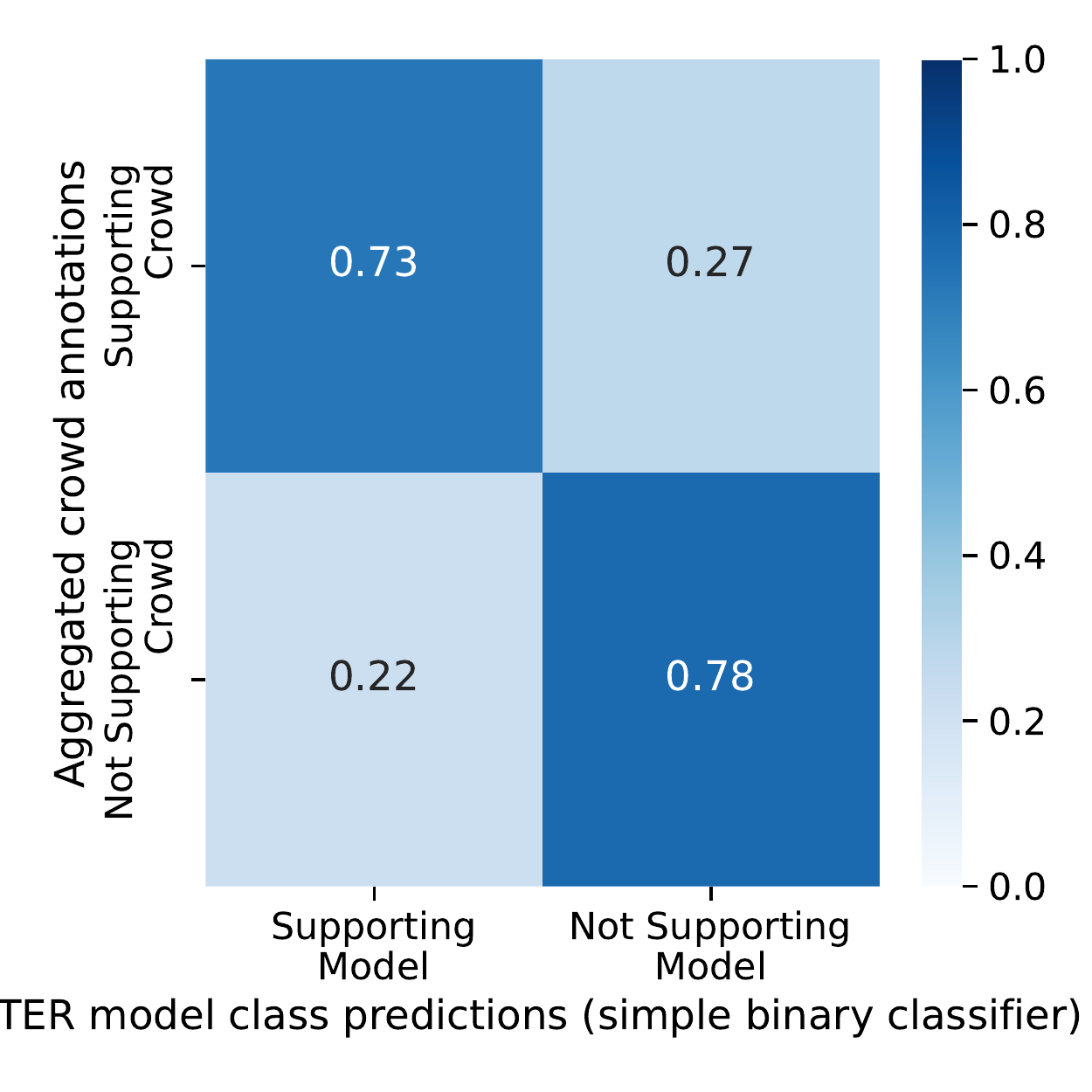}
\caption{Binary stance classes of single claim-reference pairs predicted by a RFC vs. the crowd's aggregated annotations.}
\label{fig:single_evidence_agg_CM_simple_classifier}
\end{minipage}
\hfill
\begin{minipage}{0.49\linewidth}
\includegraphics[width=1\linewidth]{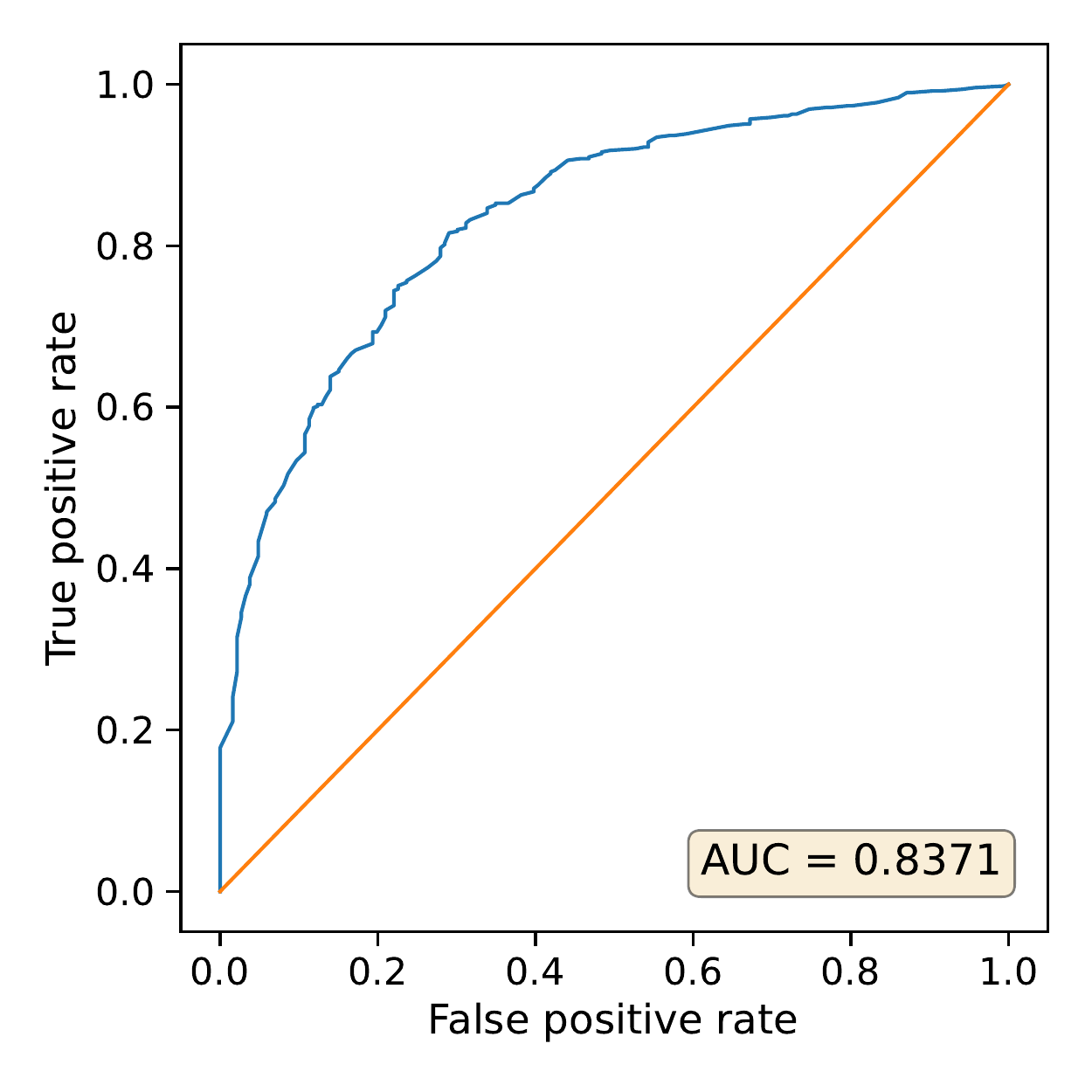}
\caption{ROC curve for the simplified binary stance classification performed by a RFC.}
\label{fig:single_evidence_agg_ROC_simple_classifier}
\end{minipage}
\end{figure}

ProVe's pipeline focuses on the classification of entire references rather than individual sentences, and the TER class probabilities are merely features for the final aggregation. Still, TER classifications for individual pieces of evidence can be greatly improved by, rather than argmax, using a simple classifier with the three TER class probabilities ($\sigma_i$), the evidence's relevance score ($\rho_i$), and the evidence's length ($|e_i|$) as features. The best scoring classifier was a Random Forests Classifier (RFC) with cross-validated (k=5) scores of $0.77$ accuracy and $0.50$ F1-score. Furthermore, by grouping the `refutes' and `neither' (NEI) classes into one `not supporting' class, turning this into a binary classification task, cross-validated (k=5) scores reach $0.79$ accuracy and $0.76$ F1-score, as well as an Area Under the ROC Curve (AUC) of $0.85$, as seen in Figures~\ref{fig:single_evidence_agg_CM_simple_classifier} and~\ref{fig:single_evidence_agg_ROC_simple_classifier}.

\paragraph{Collective TER Classification Metrics}

The annotations obtained through T2 tasks describe human judgements on the collective stance of the sets of evidence extracted from a reference towards its associated triple in WTR. After majority-voting aggregation, there are $301$ evidence sets collectively supporting the triple, $24$ refuting, and $84$ that neither support nor refute it. Taking these annotations as ground truth labels, one can measure the classification performance of ProVe. Table~\ref{tab:claim_ver_aggr_results} showcases and compares these results with different aggregation methods. Due to class imbalance, both macro averaged and weighted averaged results are reported. Not only the original ternary classification problem but also to a simplified `supporting' vs. `not supporting' binary classification problem is explored. Although a binary formulation is easier to solve, having a pipeline that provides assistance on differentiating between supporting and non-supporting references is of huge benefit to any KG's curation and editing. Results show how a simple classifier considerably outperforms the other two aggregation approaches, especially in the binary classification scenario.

\begin{table}[ht]
\centering
\begin{tabular}{lrr|rrr|rrr|}
\cline{4-9}
 &
  \multicolumn{1}{l}{} &
  \multicolumn{1}{l|}{} &
  \multicolumn{3}{l|}{Macro Averaged} &
  \multicolumn{3}{l|}{Weighted Averaged} \\ \hline
\multicolumn{1}{|l|}{Method} &
  \multicolumn{1}{l|}{Classes} &
  \multicolumn{1}{l|}{Accuracy} &
  \multicolumn{1}{l|}{Precision} &
  \multicolumn{1}{l|}{Recall} &
  \multicolumn{1}{l|}{F1} &
  \multicolumn{1}{l|}{Precision} &
  \multicolumn{1}{l|}{Recall} &
  \multicolumn{1}{l|}{F1} \\ \hline
\multicolumn{1}{|l|}{1: Weighted Sum} &
  \multicolumn{1}{r|}{3} &
  0.592 &
  \multicolumn{1}{r|}{\textbf{0.439}} &
  \multicolumn{1}{r|}{\textbf{0.484}} &
  0.439 &
  \multicolumn{1}{r|}{0.700} &
  \multicolumn{1}{r|}{0.592} &
  0.626 \\ \hline
\multicolumn{1}{|l|}{2: Malon's} &
  \multicolumn{1}{r|}{3} &
  0.641 &
  \multicolumn{1}{r|}{0.430} &
  \multicolumn{1}{r|}{0.456} &
  0.436 &
  \multicolumn{1}{r|}{0.691} &
  \multicolumn{1}{r|}{0.641} &
  0.659 \\ \hline
\multicolumn{1}{|l|}{3: RFC} &
  \multicolumn{1}{r|}{3} &
  \textbf{0.726} &
  \multicolumn{1}{r|}{0.433} &
  \multicolumn{1}{r|}{0.468} &
  \textbf{0.446} &
  \multicolumn{1}{r|}{\textbf{0.709}} &
  \multicolumn{1}{r|}{\textbf{0.726}} &
  \textbf{0.714} \\ \hline
\multicolumn{1}{|l|}{1: Weighted Sum} &
  \multicolumn{1}{r|}{2} &
  0.626 &
  \multicolumn{1}{r|}{0.609} &
  \multicolumn{1}{r|}{0.639} &
  0.596 &
  \multicolumn{1}{r|}{0.716} &
  \multicolumn{1}{r|}{0.626} &
  0.648 \\ \hline
\multicolumn{1}{|l|}{2: Malon's} &
  \multicolumn{1}{r|}{2} &
  0.667 &
  \multicolumn{1}{r|}{0.614} &
  \multicolumn{1}{r|}{0.638} &
  0.617 &
  \multicolumn{1}{r|}{0.712} &
  \multicolumn{1}{r|}{0.667} &
  0.683 \\ \hline
\multicolumn{1}{|l|}{3: RFC} &
  \multicolumn{1}{r|}{2} &
  \textbf{0.750} &
  \multicolumn{1}{r|}{\textbf{0.681}} &
  \multicolumn{1}{r|}{\textbf{0.664}} &
  \textbf{0.667} &
  \multicolumn{1}{r|}{\textbf{0.747}} &
  \multicolumn{1}{r|}{\textbf{0.750}} &
  \textbf{0.745} \\ \hline
\end{tabular}
\vspace{5pt}
\caption{Classification performance of each of the three aggregation methods on both ternary and binary collective stance TER classification formulations. Majority-voted annotations obtained in T2 are used as true labels. Results from method 3 were cross-validated with $k=5$.}
\label{tab:claim_ver_aggr_results}
\end{table}

\paragraph{Reference Representation through Retrieved Evidence}

Verifying whether ProVe properly represents entire references through the evidence set it retrieves from them is possible by comparing the evidence-level crowd annotations on the collective stance of retrieved evidence (T2 tasks) with the reference-level author annotations, which represent the stance of the reference as a whole. 

\begin{figure}[]
\begin{minipage}{0.49\linewidth}
\includegraphics[width=1\linewidth]{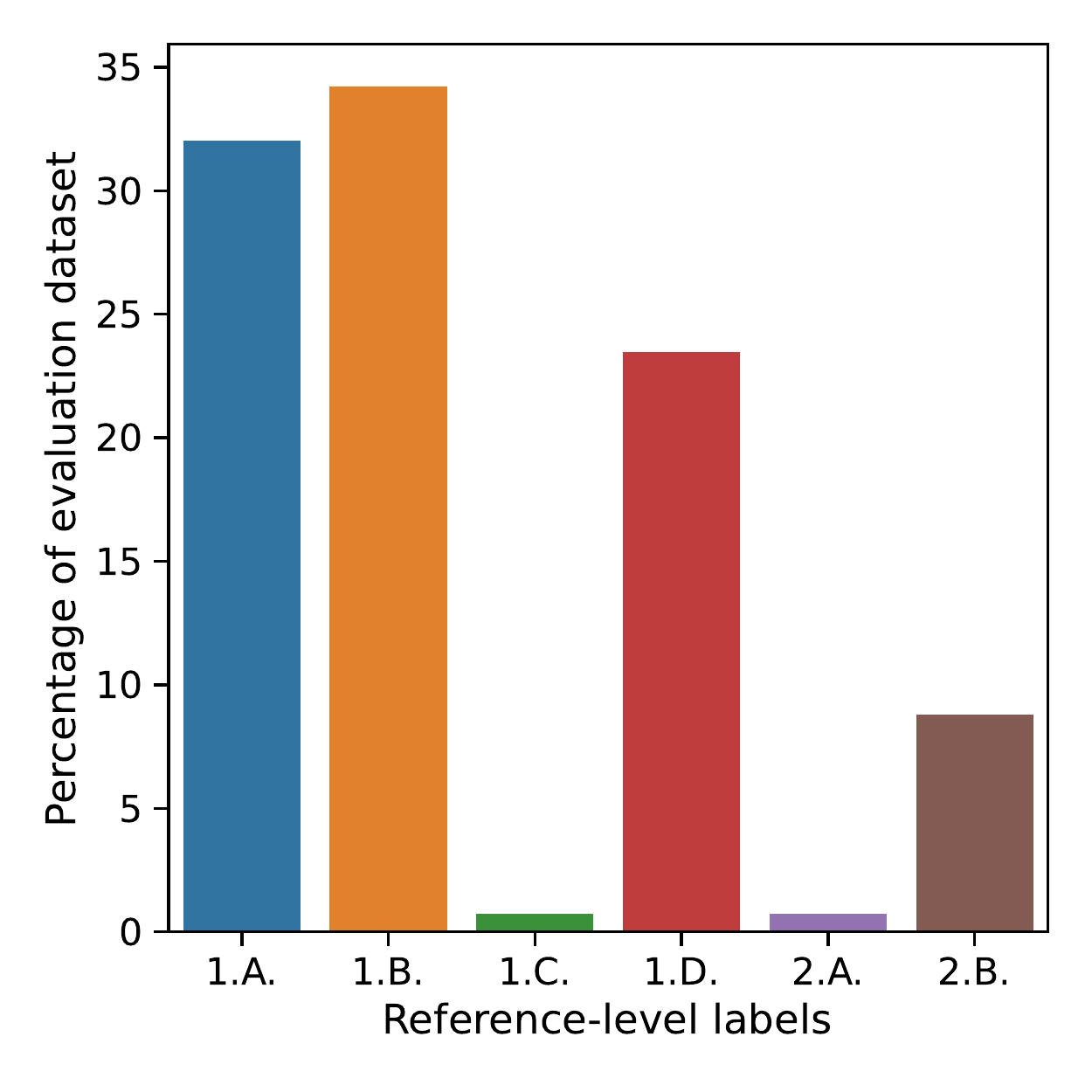}
\caption{Distribution of reference-level author labels of the evaluation dataset.}
\label{fig:manual_label_dist}
\end{minipage}
\hfill
\begin{minipage}{0.49\linewidth}
\includegraphics[width=1\linewidth]{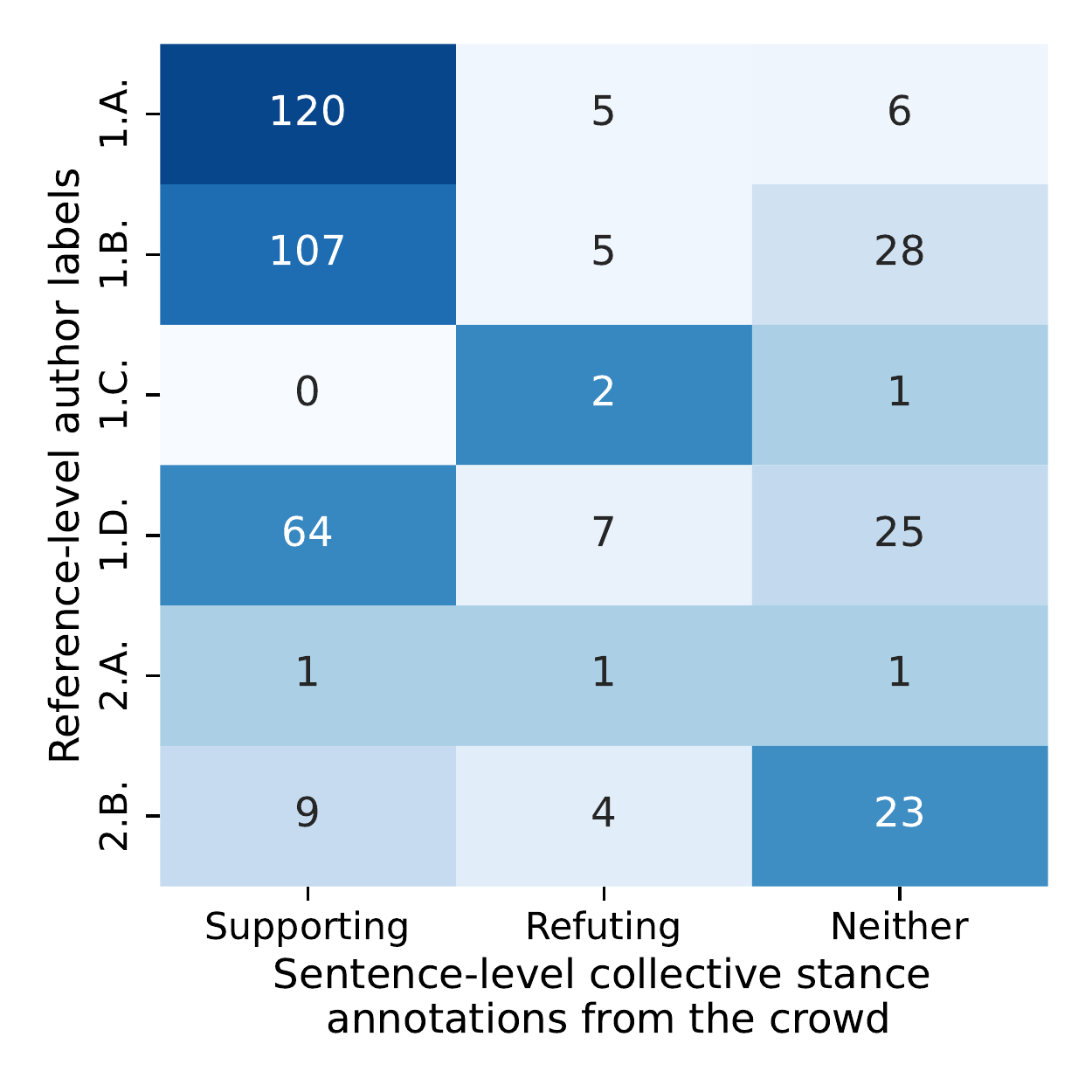}
\caption{Comparison between reference-level author label and sentence-level collective stance annotations from the crowd.}
\label{fig:manual_label_vs_crowd}
\end{minipage}
\end{figure}

As detailed in Section~\ref{sec:data_ann}, reference-level annotations consist of six categories which directly map to the three TER labels used in crowd annotations. Figure~\ref{fig:manual_label_dist} shows WTR's reference-level annotation distribution. On the ternary classification task, labels $1.A.$ through $1.D.$ map to `supporting', label $2.A.$ maps to `refuting', and label $2.B.$ to `neither'. As for the binary classification task, labels $1.X.$ map to `supporting` and labels $2.X.$ map to `not supporting`. There is a high class imbalance, with authors deeming only $2$ references as `refuting'. This number is much lower than the $24$ majority-voted by the crowd as `refuting'.

Figure~\ref{fig:manual_label_vs_crowd} shows a complete comparison between evidence-level collective stance annotations from the crowd and reference-level author labels. Crowd annotators very successfully judge references in $1.A.$ (explicit textual support in natural language), and obtain moderate to high performance on all other groups, except for $1.C.$ (non-textual support) and $2.A.$ (refuting reference). It is trivial to see why ProVe, a text-based approach, fails at $1.C$. The failure at $2.A.$ is partially due to the disparity between refuting text seen in training and in the evaluation dataset, as explained in Section~\ref{sec:disc_ref}. The low amount of refuting references compromises a deeper look. Still, where supportive textual information is the most available, explicit, and naturally written, the better ProVe performs.

\paragraph{Full Pipeline Performance}

Lastly, ProVe's performance on a `supported' vs. `not supported' binary classification scenario is investigated by comparing the outputs obtained with the WTR evaluation dataset to its reference-level annotations provided by the authors. The variation of such performance based on how supportive information is expressed on text is also analysed.

ProVe is evaluated on the entire WTR by using the reference-level annotations as ground-truth and adopting ProVe's best performing aggregation method, the simple classifier (method $\#3$). Additionally, for each of the supporting reference-level labels $1.A.$ through $1.D.$ (except for $1.C$, as it is non-textual), WTR is modified by keeping only those triple-reference pairs with either that specific supporting label or labelled as `not supporting'. ProVe is then also evaluated on these modified datasets. Due to the very low amount of $2.A.$ labels, only the binary `supporting' vs. `not supporting' classification task is evaluated. Table~\ref{tab:rfc_results_per_author_label} showcases the results obtained. It shows that ProVe has a good result on the evaluation dataset overall, with close to $80\%$ accuracy, and an excellent performance on identifying support from references that showcase said support through explicit and naturally written textual information ($1.A.$). It also shows good results on references where support is not naturally written ($1.B.$).  

\begin{table}[ht]
\begin{tabular}{lr|rrr|rrr|r}
\cline{3-8}
 &
  \multicolumn{1}{l|}{} &
  \multicolumn{3}{l|}{Macro Averaged} &
  \multicolumn{3}{l|}{Weighted Averaged} &
  \multicolumn{1}{l}{} \\ \hline
\multicolumn{1}{|l|}{Support type} &
  \multicolumn{1}{l|}{Accuracy} &
  \multicolumn{1}{l|}{Precision} &
  \multicolumn{1}{l|}{Recall} &
  \multicolumn{1}{l|}{F1-Score} &
  \multicolumn{1}{l|}{Precision} &
  \multicolumn{1}{l|}{Recall} &
  \multicolumn{1}{l|}{F1-Score} &
  \multicolumn{1}{l|}{AUC} \\ \hline
\multicolumn{1}{|l|}{1.A.} &
  \textbf{0.875} &
  \multicolumn{1}{r|}{\textbf{0.821}} &
  \multicolumn{1}{r|}{\textbf{0.838}} &
  \textbf{0.829} &
  \multicolumn{1}{r|}{\textbf{0.878}} &
  \multicolumn{1}{r|}{\textbf{0.875}} &
  \textbf{0.876} &
  \multicolumn{1}{r|}{\textbf{0.908}} \\ \hline
\multicolumn{1}{|l|}{1.B.} &
  0.779 &
  \multicolumn{1}{r|}{0.668} &
  \multicolumn{1}{r|}{0.637} &
  0.648 &
  \multicolumn{1}{r|}{0.762} &
  \multicolumn{1}{r|}{0.779} &
  0.768 &
  \multicolumn{1}{r|}{0.745} \\ \hline
\multicolumn{1}{|l|}{1.D.} &
  0.666 &
  \multicolumn{1}{r|}{0.662} &
  \multicolumn{1}{r|}{0.669} &
  0.649 &
  \multicolumn{1}{r|}{0.749} &
  \multicolumn{1}{r|}{0.666} &
  0.682 &
  \multicolumn{1}{r|}{0.754} \\ \hline
\multicolumn{1}{|l|}{ALL} &
  0.794 &
  \multicolumn{1}{r|}{0.565} &
  \multicolumn{1}{r|}{0.635} &
  0.574 &
  \multicolumn{1}{r|}{0.864} &
  \multicolumn{1}{r|}{0.794} &
  0.821 &
  \multicolumn{1}{r|}{0.753} \\ \hline
\end{tabular}
\vspace{5pt}
\caption{ProVe's binary classification performance on all WTR and per type of textual support. Reference-level annotations were used as ground-truth, and a simple classifier as aggregation method. Values obtained through cross-validation ($k=5$).}
\label{tab:rfc_results_per_author_label}
\end{table}
\section{Discussions and Conclusions}
\label{sec:disc} % 1.5 pages

In this section, aspects and limitations of the implementation and evaluation results of ProVe are further discussed. Additionally, future directions of research are pointed out and final conclusions are drawn.

\subsection{ProVe for Fact Verification}

Fact checking as a tool to assist users in discerning between factual and non-factual information has a myriad of applications, formulations, approaches, and, overall, considerably ambiguous results. The effects of fact-checking interventions, while significant, are substantially weakened by its targets' preexisting beliefs and knowledge~\cite{walter2020fact}. Its effectiveness depend heavily on many variables, such as the type of scale used and whether facts can be partially checked, as well as different impacts if they go along or against a person's ideology. This further motivates ProVe's standpoint of judging support instead of veracity. Triples are evaluated not as factual or non-factual, but based on their documented provenance, passing the onus of providing trustworthy and authoritative sources to the graph's curators. This keeps ProVe's judgements from clashing with the ideologies of its users, as the pipeline does not pass factual, but linguistic judgement. Additionally, by using only two to three levels of verdict and not including graphical elements, the presence of elements that compromise fact-checking~\cite{walter2020fact} is hampered. The authors' focus for future research lies in increasing ProVe's explainability in order to increase trust and understanding.

The results achieved by ProVe, especially on text-rich references, are considered by the authors as more than satisfactory, representing an excellent addition to a family of approaches that currently includes very few, e.g. DeFacto~\cite{gerber2015defacto} and FactCheck~\cite{syed2018factcheck}. Still, there is a need for a dataset specialised in AFC on KGs in order to tie in these approaches and make benchmarking them and future works possible. While WTR (ProVe's evaluation dataset) can serve this purpose, it can definitely be improved in size and predicate coverage.

ProVe's use as a tool can greatly benefit from an active learning scenario which would further enhance the models and techniques it employs. As the same time, users of the tool inherently introduce a bias based on their demographics, with the same being valid for the crowdsourced evaluation of ProVe's pipeline. Being aware of this bias is crucial to the proper deployment of such approaches. 

\subsection{Text Extraction Add-ons}

ProVe's text extraction module has three essential steps: rendering web pages with a web crawler, using rule-based methods to convert content inside HTML tags into text, and sentence segmentation. While the rule-based methods presented in this paper are simple, they are quite effective, as shown in Section~\ref{sec:eval}. Better and more specialized rules and methods to detect and extract text from specific HTML layouts, such as turning tabular structures or sparse infoboxes into sequential and syntactically correct sentences, can be seamlessly integrated into ProVe. Both supervised and unsupervised approaches can also be applied. 

In order to properly assess such added methods, as well as to provide more insight into the text extraction module in general, a direct evaluation of its performance would be extremely helpful. Although good performance on downstream tasks is a good indicator, it does not indicate where to improve text extraction. Be it through descriptive statistics or comparison against golden data, this as a focus of future research alongside model explainability

\subsection{Usage of Qualifiers}
\label{sec:disc_quali}

Triples in KGs such as Wikidata often are accompanied by qualifiers that further detail it. A triple such as the one seen in Figure~\ref{fig:example} ($<$\textit{Librarian of Congress, position holder, James H. Billington}$>$) has several qualifiers, such as `start time' and `end time'. If a person were Librarian of Congress twice, this qualifier would differentiate two triples that would otherwise be identical if expressed only with main components. ProVe does not take qualifiers into consideration. As such, two triples with distinct IDs and meanings can have exactly the same verbalisations which, while adequate, do not contain all information.

Ribeiro et al.~\cite{ribeiro2020investigating} show transformers can verbalise multiple triples into a single sentence. Hence, adding qualifiers as secondary triples is possible, generating more detailed verbalisations. However, ProVe's sentence selection and claim verification modules contain models fine-tuned on FEVER, whose vast majority of sentences contain only a main piece of information with little to no additional details, e.g. `Adrienne Bailon is an accountant' and `The Levant was ruled by the House of Lusignan'. In order to make proper use of qualifiers during verbalisation, there needs to be an assurance that downstream modules can properly handle more complex sentences by either different or augmented training data.
    
\subsection{Detecting Refuting Sources with FEVER}
\label{sec:disc_ref}

The FEVER dataset presents claims that are normally short and direct in nature, from multiple domains, and associated evidence extracted directly from Wikipedia. ProVe shows it is possible to use FEVER to train pipeline modules to detect supportive and non-supportive sources evidence. However, as seen in Section~\ref{sec:eval_claimverification}, detecting refuting sources is hard for ProVe and believed to be due to how FEVER generates refuted claims through artificial alterations. Claims labelled by FEVER as `REFUTES' are those generated by annotators who alter claims that would otherwise be supported by its associated evidence. Alterations follow six types: paraphrasing, negation, entity/relationship substitution, and making the claim more general/specific. This leads to claims that, while meaningful and properly annotated, would never be encoded in a KG triple, such as ``As the Vietnam War raged in 1969, Yoko Ono and her husband John Lennon \textbf{did not} have two week-long Bed-Ins for Peace'' or ``Ruth Negga \textbf{only} acts in Irish cinema''. Additionally, associated evidence often rely on common sense in order to refute these claims, such as ``Kingdom Hearts III is owned by Boyz II Men'', whose relevant evidence at no point elaborate on Kingdom Hearts III's ownership, only describing it as a Japanese video-game. We supposedly know Boyz II Men is a music group, rendering the claim implausible.

While useful for other tasks, these refuted claims are very different from refutable triples occurring naturally in KGs, which mainly consist of triples whose objects have different values in the provenance. One such example is ``Robert Brunton was born on 23/03/1796'', whose reference actually mentions the ``10th of February 1796''. In order to properly detect KG provenance that refute their triples, ProVe's claim verification module needs re-training on a fitting subset of FEVER, or on a new dataset containing non-artificial refuted claims.

\subsection{Conclusions}
Knowledge graphs are widespread secondary sources of information. Their data is extremely useful and available in a semantic format that covers a myriad of domains. Ensuring the verifiability of this data through documented provenance is a task that is crucial to the upkeep of their usability and one that should be actively supported by automated and semi-automated tools to help data curators and editors cope with the sheer volume of information. However, as of now, there are no such tools deployed at large scale KGs and only a very small family of approaches tackle this task from a research standpoint.

This paper proposes, describes, and evaluates ProVe, a pipelined approach to support the upkeep of KG triple verifiability through their documented provenance. ProVe leverages large pre-trained LMs, rule-based methods, and simple classifiers, to provide automated assistance to the activity of creating and maintaining references in a KG. ProVe's pipeline aims at extracting relevant textual information from references and evaluating whether or not they support an associated KG triple, providing its users with a support classification, a support probability, as well as relevance and textual entailment metrics for the evidence used. Deployed correctly, ProVe can help detect verifiability issues in existing references, as well as improve the reuse of good sources. Additionally, the approach can be expanded to work in a multilingual setting.

ProVe has been evaluated with WTR, a dataset of triple-reference pairs extracted directly from Wikidata, a large KG, and annotated by both crowdworkers and the authors. ProVe achieves $75\%$ accuracy, $0.681$ F1-macro, and $0.667$ AUC on the full evaluation dataset, which includes references to many different web domains. On references where support is stated explicitly and in natural text, ProVe achieves an excellent $87.5\%$ accuracy ($0.829$ F1-macro and $0.908$ AUC).

Future work mainly lies in exploring techniques to improve ProVe's explainability, with a focus on its sentence selection and claim verification steps. Other directions can include expanding the size and the distinct predicate coverage of the benchmarking dataset WTR, as well as a direct evaluation of text extraction and segmentation techniques.

%Additionally, WTR can be used as a benchmarking dataset for future approaches to AFC on KGs on textual references from multiple types of web domains, not only Wikipedia. Our crowdsourcing tasks, whose code and implementation is available on GitHub, can also be reused for similar studies.

\paragraph*{Acknowledgements} This research received funding from the European Union’s Horizon 2020 research and innovation programme under the Marie Skłodowska-Curie grant agreement no. 812997.

\bibliographystyle{splncs04}           % Style BST file.
\bibliography{bibliography}        % Bibliography file (usually '*.bib')

% or include bibliography directly:
%\begin{thebibliography}{0}
%\bibitem{r1} F. Author, Information about cited object.
%
%\bibitem{r2} S. Author and T. Author, Information about cited object.
%\end{thebibliography}

\newpage
\appendix
\section{Crowdsourcing Task Designs}
\label{appendix:1}

\begin{figure}[ht]
  \centering
  %\vspace*{-15pt}
  \includegraphics[width=0.9\linewidth]{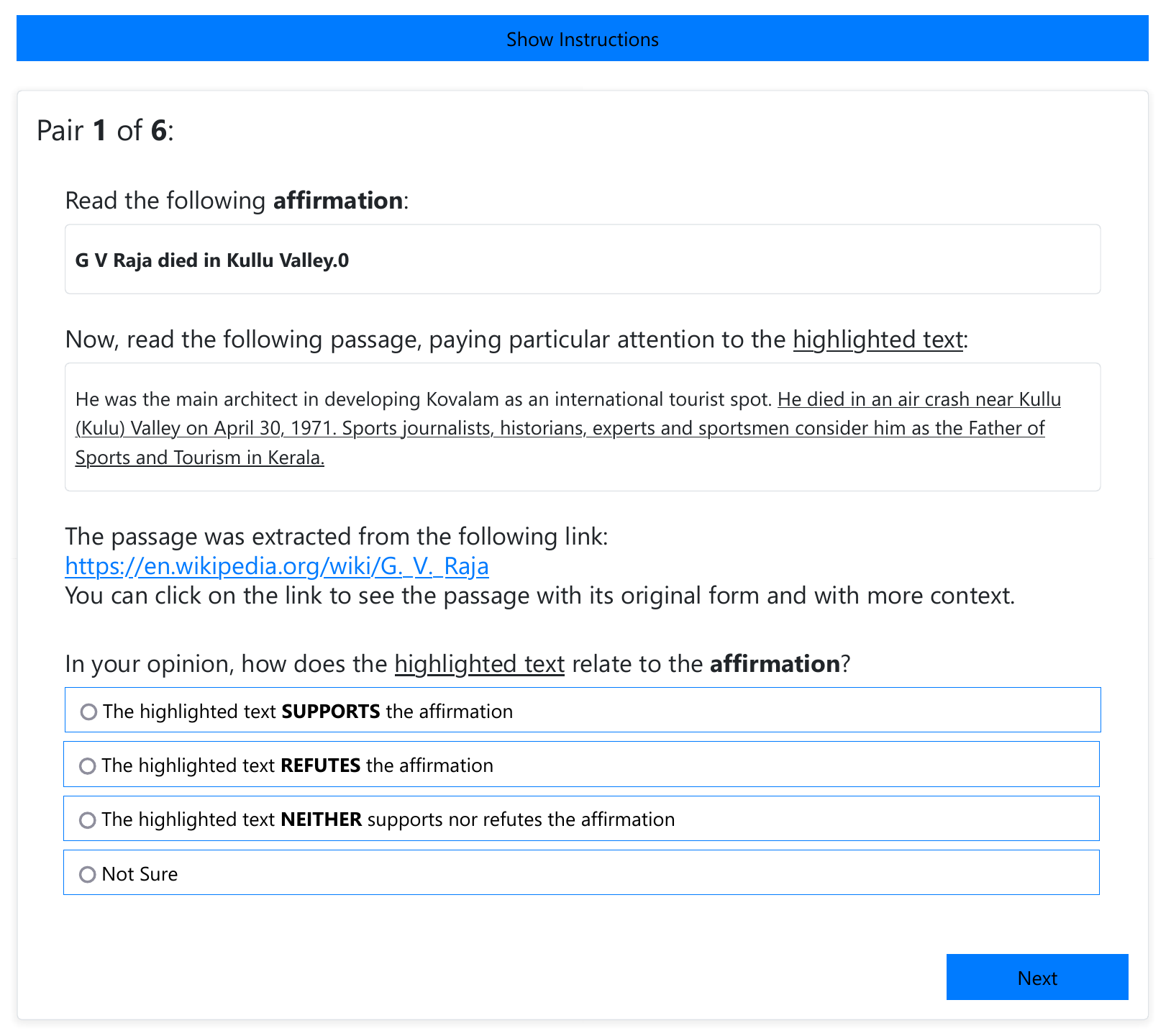}
  %\vspace*{-15pt}
  \caption{The task design T1, which collects evidence-level annotations of the individual stances of pieces of evidence towards a Wikidata triple.}
  \label{fig:T1}
%\vspace*{-15pt}
\end{figure}

\begin{figure}[]
  \centering
  %\vspace*{-15pt}
  \includegraphics[width=0.9\linewidth]{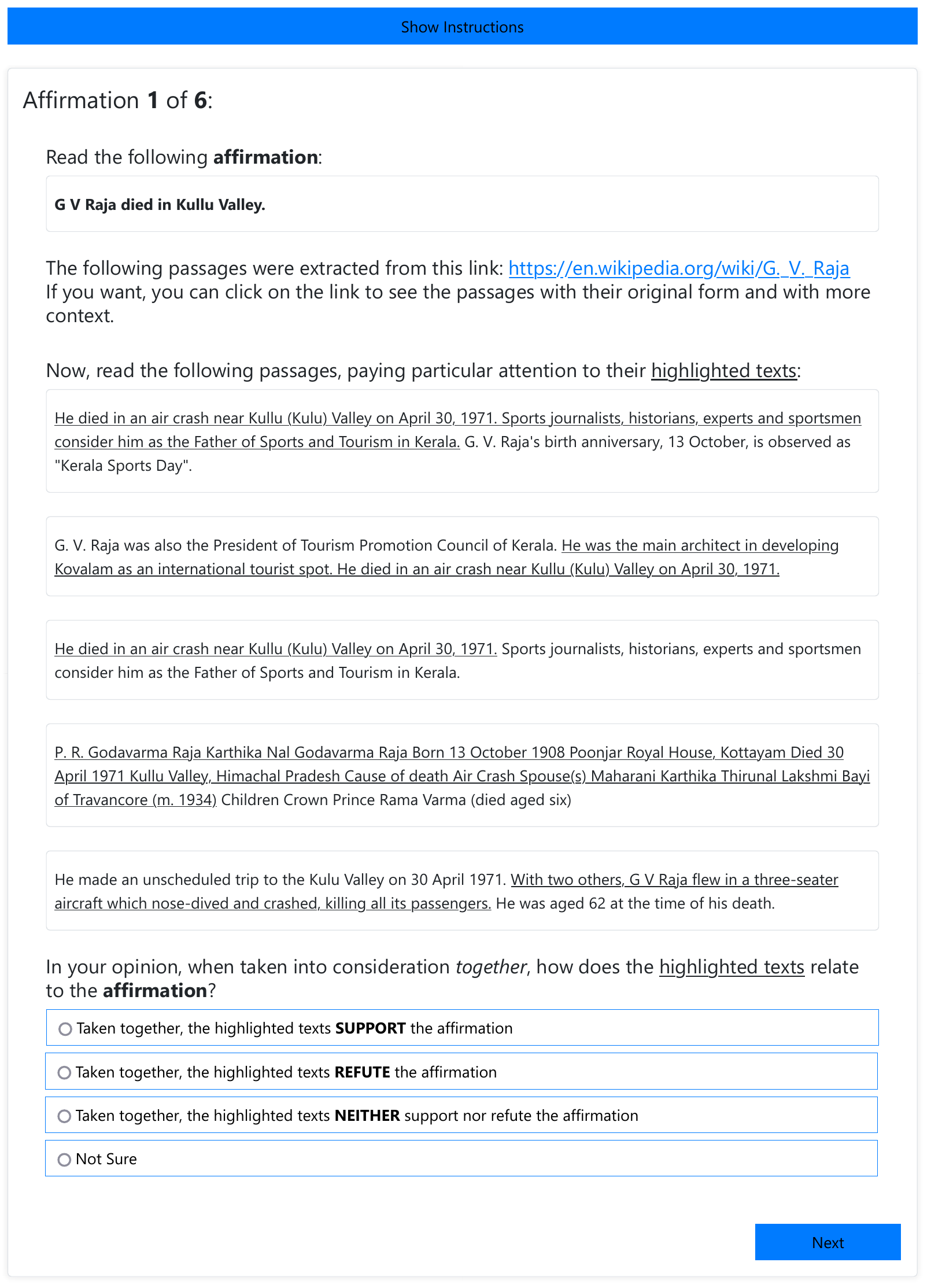}
  %\vspace*{-15pt}
  \caption{The task design T2, which collects evidence-level annotations of the collective stances of evidence sets towards a Wikidata triple.}
  \label{fig:T2}
%\vspace*{-15pt}
\end{figure}
\newpage
\section{WTR Dataset Format}
\label{appendix:2}

WTR is available at Figshare~\footnote{\url{https://figshare.com/s/df0ec1c233ebd50817f4}} and contains $409$ Wikidata triple-reference pairs, representing $32$ groups of text-rich web domains commonly used as sources, as well as $76$ distinct Wikidata properties. Out of the $416$ sampled triple-reference pairs, as described in Section~\ref{sec:data_con}, $7$ were left out due to having exactly the same triple components and referenced URL as other triple-reference pair, as explained in Section~\ref{sec:disc_quali}. $43\%$ of references were obtained through external IDs and $57\%$ through direct URLs. Each entry has the following attributes:

\begin{itemize}
    \item Reference attributes:
    \begin{itemize}
        \item Reference ID: An unique identifier issued to the reference by Wikidata;
        \item Reference property ID: The unique identifier of the Wikidata property used by the reference to encode the URL we retrieve for it;
        \item Reference datatype: Whether the reference's URL was retrieved by a direct URL or an URL formatted with an external identifier;
        \item URL: The URL retrieved for this reference;
        \item Netloc: The actual web domain of this URL;
        \item Netloc group: The web domain of this URL after grouping references under the RARE and OTHER groups;
        \item Final URL: The URL reached after redirects and whose HTML and text were extracted by ProVe into sentences for annotation;
        \item HTML: The HTML code extracted from the reference's final URL.
    \end{itemize}
    \item Claim attributes:
    \begin{itemize}
        \item Claim ID: An unique identifier issued to the claim by Wikidata;
        \item Rank: The claims's rank, either normal or preferred;
        \item Datatype: The datatype of the claim's object, e.g. quantity, string, etc;
        \item Component IDs: The Wikidata IDs of the claim's subject and property;
        \item Component labels: Main labels for subject, property, and object;
        \item Component aliases: Aliases lists for subject, property, and object;
        \item Component descriptions: Wikidata descriptions for subject, property, and object (if it is a Wikidata item).
    \end{itemize}
    \item Annotations for evaluation:
    \begin{itemize}
        \item Evidence-level annotations (T1): The evidence-level annotations that describe the individual TER stances of each piece of evidence towards the claim, in which the evidence set is the five most relevant passages collected from the URL. This consists the following attributes for each piece of evidence:
        \begin{itemize}
    		\item Evidence: The individual textual evidence collected and being annotated;
    		\item MTurk IDs: A list of anonymous worker IDs and assignment IDs denoting the crowd workers who provided annotations/votes;
    		\item TER Relation: The list of TER stances voted by the workers, where 0 = SUPP, 1 = REF, 2 = NEI, 3 = Not Sure;
    		\item Reason for 'Not Sure;: If a voter gave 'not sure' as their relation, this denotes the reason why, out of the list of options seen in this paper's appendix or a free-text reason;
    		\item Times: The times in seconds taken by a worker to provide their full annotations;
    		\item Aggregated TER Relation: The majority-voting aggregation of each individual worker's TER stance annotation;
	    \end{itemize}
	    \item Evidence-level annotations (T2): The evidence-level annotations that describe the collective TER stances of the entire evidence set towards the claim, in which the evidence set is the five most relevant passages collected from the URL. This consists of the following attributes: 
	    \begin{itemize}
    		\item Evidence: The entire textual evidence set collected and being collectively annotated;
    		\item MTurk IDs: A list of anonymous worker IDs and assignment IDs denoting the crowd workers who provided annotations/votes;
    		\item TER Relation: The list of TER stances voted by the workers, where 0 = SUPP, 1 = REF, 2 = NEI, 3 = Not Sure;
    		\item Reason for 'Not Sure;: If a voter gave 'not sure' as their relation, this denotes the reason why, out of the list of options seen in this paper's appendix or a free-text reason;
    		\item Times: The times in seconds taken by a worker to provide their full annotations;
    		\item Aggregated TER Relation: The majority-voting aggregation of each individual worker's TER stance annotation;
	    \end{itemize}
	    \item Sentence-level author annotations: The sentence-level annotation representing the stance of the entire reference towards the triple.
    \end{itemize}
\end{itemize}

\end{document}